\documentclass[journal]{IEEEtran}

\usepackage{amsfonts}
\usepackage{color}
\usepackage{bm}
\usepackage{mathrsfs}
\usepackage{multirow}
\usepackage{balance}
\usepackage{url}

\usepackage{subfigure}

\usepackage[utf8]{inputenc}
\usepackage[english]{babel}
\usepackage{amsmath, amssymb}
\usepackage{float}
\usepackage{makecell}
\usepackage{multirow}
\usepackage{cite}

\usepackage{amsmath,amssymb,amsfonts}
\usepackage{amsmath}
\usepackage{amsfonts}
\usepackage{amssymb}\usepackage{graphicx}
\usepackage{textcomp}
\usepackage{xcolor}

\usepackage{subfigure}
\usepackage{epstopdf}
\usepackage{amsmath}
\usepackage{algorithm}
\usepackage{algorithmic}

\ifCLASSOPTIONcompsoc
 \usepackage[caption=false,font=normalsize,labelfont=sf,textfont=sf]{subfig}
\else
 \usepackage[caption=false,font=footnotesize]{subfig}
\fi

\usepackage{stfloats}
\begin{document}

%
\title{Spherical Interpolated Convolutional Network with Distance-Feature Density for 3D Semantic Segmentation of Point Clouds}
%
%
%

\author{Guangming Wang, Yehui Yang, Huixin Zhang, Zhe Liu, and Hesheng Wang
 \thanks{This work was supported in part by the Natural Science Foundation of China under Grant U1613218, 61722309 and U1913204, in part by Beijing Advanced Innovation Center for Intelligent Robots and Systems under Grant 2019IRS01. Corresponding Author: Zhe Liu and Hesheng Wang. The first two authors contributed equally.}
 \thanks{G. Wang, Y. Yang, H. Zhang, and H. Wang are with Department of Automation, Insititue of Medical Robotics, Key Laboratory of System Control and Information Processing of Ministry of Education, Key Laboratory of Marine Intelligent Equipment and System of Ministry of Education, Shanghai Jiao Tong University, Shanghai 200240, China. H. Wang is with Beijing Advanced Innovation Center for Intelligent Robots and Systems, Beijing Institute of Technology, China. Z. Liu is with the Department of Computer Science and Technology, University of Cambridge.}

}

%
%

\markboth{Journal of \LaTeX\ Class Files,~Vol.~14, No.~8, August~2015}%
{Shell \MakeLowercase{\textit{et al.}}: Bare Demo of IEEEtran.cls for IEEE Journals}
%



\maketitle

\begin{abstract}
The semantic segmentation of point clouds is an important part of the environment perception for robots. However, it is difficult to directly adopt the traditional 3D convolution kernel to extract features from raw 3D point clouds because of the unstructured property of point clouds. In this paper, a spherical interpolated convolution operator is proposed to replace the traditional grid-shaped 3D convolution operator. This newly proposed feature extraction operator improves the accuracy of the network and reduces the parameters of the network. In addition, this paper analyzes the defect of point cloud interpolation methods based on the distance as the interpolation weight and proposes the self-learned distance-feature density by combining the distance and the feature correlation. The proposed method makes the feature extraction of spherical interpolated convolution network more rational and effective. The effectiveness of the proposed network is demonstrated on the 3D semantic segmentation task of point clouds. Experiments show that the proposed method achieves good performance on the ScanNet dataset and Paris-Lille-3D dataset.
\end{abstract}


%
\IEEEpeerreviewmaketitle

\vspace{-3mm}
\section{Introduction}

\IEEEPARstart{B}{oth} the RGB-D camera and LiDAR can obtain dense point clouds. The semantic segmentation of the point cloud is of great significance to the 3D environment perception for service robots and autonomous driving~\cite{mei2019semantic,xiang2019novel}. However, learning from point clouds is still not effective enough. How to apply the research achievements of 2D images~\cite{qiao2013introducing,liu2013gnccp} to the point clouds is a great challenge brought by the unstructured property of point clouds. In recent years, there are many methods proposed to deal with this challenge, including the methods by querying and selecting the fixed number of points in the neighborhood and performing the Multi-Layer Perceptrons (MLP)~\cite{qi2017pointnet,qi2017pointnet++}, methods of graph network combining the mutual graph relationship between points~\cite{wang2019dynamic}, methods of point-wise convolution~\cite{hua2018pointwise}, etc. Most of these methods perform feature extraction from each local neighborhood based on K Nearest Neighbors (KNN) or ball query~\cite{qi2017pointnet++}. These two extraction methods contain defects. The KNN obtains a small region in the dense area, and a large region in the sparse area~\cite{hermosilla2018monte}. This method of taking different scales for different point densities makes the same feature extraction operator cover different scales of area. In addition, the classic ball query~\cite{qi2017pointnet++} fixes the scale of the area, but the way of fixing the number of points within the sphere makes it to copy points when points are not enough or abandon further points when the number of points exceeds its demands. These all increase uncertainty and difficulty in the learning process. InterpConv~\cite{mao2019interpolated} proposed to interpolate point clouds in the space with a certain weight into the eight corners of a cube in which points is located, thus realizing the ordering of point clouds. However, for each convolution kernel weight, its receptive field is not spatially symmetric. That is, points with the same distance to a corner of the cube cannot be completely interpolated to this corner, which is unreasonable for extracting the spatial feature of the corner.

\begin{figure}[t]
	\begin{center}
		\includegraphics[width=1.0\linewidth]{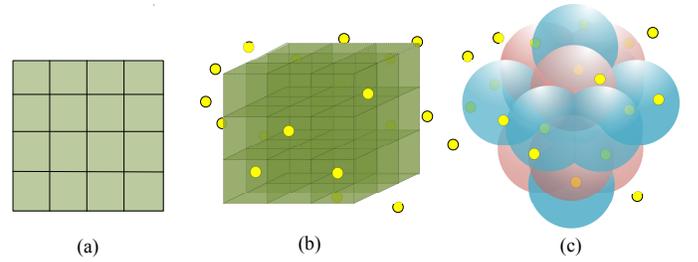}
	\end{center}
	\vspace{-4mm}
	\caption{Comparison of 2D convolution for images and two types of 3D convolutions for point clouds. (a) shows the 2D image convolution. (b) shows the cube-shaped convolution of point clouds. (c) shows the proposed spherical convolution for point clouds.}
	\label{fig1}
\end{figure}
\par As shown in Fig.~\ref{fig1}, images can be regarded as a matrix which makes it simple for the 2D convolution kernel to be used repeatedly and cover all pixels intensively. However, for unstructured point clouds, 3D grid-shaped convolution kernel is no longer natural and brings more learning parameters. Since the sphere is the perfect symmetrical structure, designing an effective convolution kernel based on the symmetrical structure of the sphere may be an important aspect to improve the extraction ability of the convolution.
\par Inspired by the densely packed form of metal, which is proved to have the highest density that can be achieved by any arrangement of spheres~\cite{hales1998overview,szpiro2003mathematics,hales2017formal}, an extensible spherical-based dense spatial feature extraction operator is designed in this paper. The comparative experiment proves that with the same network structure, the spherical interpolated convolution operator uses fewer learning parameters but brings more benefits than grid-shaped convolution, which shows the effectiveness of dense spherical interpolated convolution in 3D space.

\par Although the calculation of feature in InterpConv~\cite{mao2019interpolated} is free from the density of point clouds due to normalization, it is still affected by feature density. In one spherical cell, the feature-intensive area will have a greater impact on the feature of the cell center and repeated point features in a local region can lead to a wrong extracted feature. Therefore, the concept of distance-feature density is proposed, so that the calculation of spatial feature will not be affected by the density of points and the uneven density of features.
\par The main contributions of this paper are as follows: 
\begin{itemize}
	\item  A dense spherical interpolated convolution operator is proposed for the unstructured point clouds in 3D space. The fewer network parameters are realized by the novel operator than the 3D grid-shaped convolution operator under the same network structure so that it can reduce memory usage.
	\item  According to the characteristics of spatial feature calculation, the concept of distance-feature density is proposed. By effectively learning and combining distance-feature density to spatial feature calculation, the calculation of spatial feature is more reasonable and effective.
	\item  The semantic segmentation network is designed based on the proposed dense spherical interpolated convolution operator with distance-feature density. The experiment results show that the spherical interpolated convolution network has less learning parameters than the grid-shaped convolution network, but achieves better performance. Distance-feature density also improves the experimental results. The method achieves competitive performance on ScanNet dataset~\cite{dai2017scannet} and Paris-Lille-3D dataset~\cite{roynard2018paris}.
\end{itemize}

The organization of the paper is as follows: Section II outlines the related work to semantic segmentation of point cloud. Section III describes the proposed spherical interpolated convolutional network and the distance-feature density. Section IV provides our experiments and discussions on the proposed network. Finally, Section V draws a conclusion.

\section{Related Work}\label{related}

\subsection{Multi-view Representation}

In multi-view based methods, 2D convolution is performed after projecting 3D data to 2D~\cite{su2015multi,boulch2017unstructured,8584494,tatarchenko2018tangent}. Su et al.~\cite{su2015multi} first proposed multi-view CNN for 3D shape recognition. Boulch et al.~\cite{boulch2017unstructured} selected various camera positions to generate virtual images of RGB and depth for semantic segmentation of point clouds. Tatarchenko et al.~\cite{tatarchenko2018tangent} proposed the method of projecting a local neighborhood onto a local tangent plane for semantic segmentation.

\subsection{Volumetric Representation}

3D CNN is applied on volumetric representation to project point clouds onto a 3D grid of Euclidean space~\cite{ben20183dmfv,8287819,9072337}. However, voxel-based methods suffer from low efficiency in computation and storage, and the performance is limited by voxel resolution. Combining the sparseness of the point clouds in the voxel mapping process, OctTree structure~\cite{riegler2017octnet,wang2017cnn} or hash-map~\cite{graham20183d} can achieve higher resolution and enhanced performance.

\subsection{Point Cloud Representation}

LiDAR and depth cameras commonly used in autonomous robots both generate raw data as point clouds. Deep learning-based approaches on point clouds can directly learn from the raw output of sensors.

\subsubsection{MLP for point clouds} PointNet~\cite{qi2017pointnet} is considered to be a pioneer in deep learning on point clouds. Later, PointNet++~\cite{qi2017pointnet++} and So-Net~\cite{li2018so} started to develop hierarchical structures like the classic convolutional network~\cite{simonyan2014very,he2016deep}. But they all used MLP to aggregate local neighborhood information. Although the convolution kernel can be implemented by MLP~\cite{li2018pointcnn,wang2018deep,hermosilla2018monte} as MLP can approximate any continuous function, the use of such representations makes the convolution operator more complex and makes the network convergence more difficult.

\subsubsection{Graph CNN on point clouds} Recently, graph convolution is also applied to solve the learning problem of disordered point clouds~\cite{wang2019dynamic,liang2019hierarchical,landrieu2018large,landrieu2019point,wang2019graph,jiang2019hierarchical}. Different from that they wish to consider the spatial correlation of points, the proposed method in this paper applies convolution on Euclidean space. In our proposed method, efficient ways of feature extraction are desired to adopt the convolution more rationally and directly.

\begin{figure*}[t]
	\begin{center}
		\vspace{0mm}
		\includegraphics[width=0.95\linewidth]{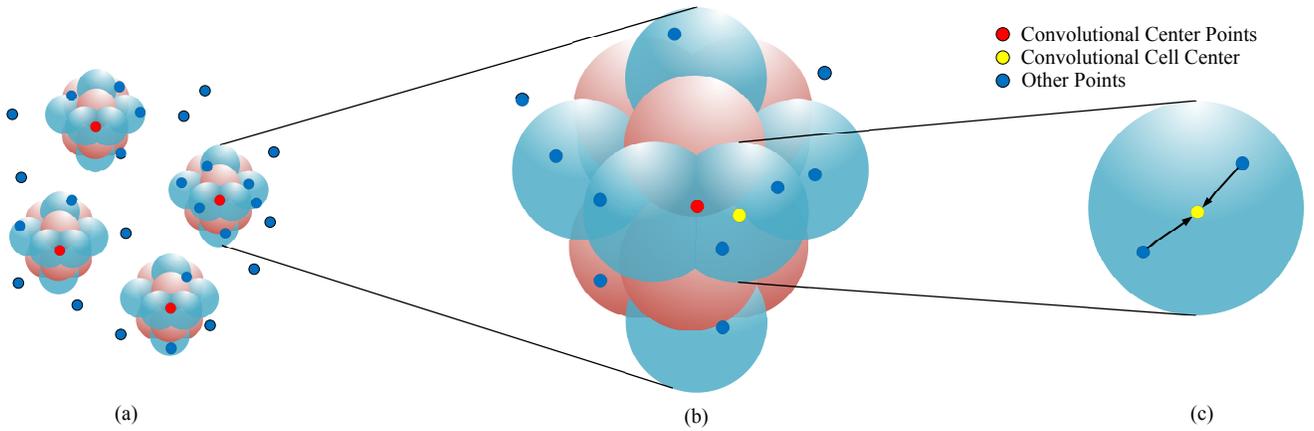}
	\end{center}
	\vspace{-4mm}
	\caption{Schematic diagram of spherical interpolated convolution. (a) shows the center points of each convolution kernel obtained by farthest point sampling. (b) shows the internals of a convolution kernel and the schematic diagram that a spherical cell extract features from point clouds. (c) shows a schematic diagram of using neighborhood points around a cell to get the feature of the cell center.}
	\label{fig6}
\end{figure*}

\subsubsection{Convolutional networks for point clouds} Some recent works also define convolution kernels for point clouds. Pointwise~\cite{hua2018pointwise} has no hierarchical structure, and points in a local area with different locations have the same convolution weights. SPLATNet~\cite{su2018splatnet} projected the input features of point clouds onto the high-dimension mesh, and then used the bilateral convolution to extract features.  SpiderCNN~\cite{xu2018spidercnn} has different weights for each neighbor of the convolution kernel center. PointConv~\cite{wu2019pointconv} proposed an MLP network to learn weight function and proposed an inverse density module to improve semantic segmentation effect. Komarichev et al.~\cite{komarichev2019cnn} proposed annular convolution by considering ring-shaped structures and directions before the regular convolution operation. Lei et al.~\cite{lei2019octree} proposed spherical kernels for point clouds, which uses a whole spherical kernel to partition the space and coarsens the representation of sparse point clouds. Zhao et al.~\cite{zhao2019pointweb} used an Adaptive Feature Adjustment (AFA) module to densely connect every two points in a local neighborhood for better representing than DGCNN~\cite{wang2019dynamic}. Zhang et al.~\cite{zhang2019shellnet} carried out the max pooling over all points in each concentric spherical shell and also designed the ShellNet for object classification, part segmentation, and semantic segmentation. Mao et al.~\cite{mao2019interpolated} proposed interpolated convolution for 3D point clouds, which mostly conforms to the traditional 2D convolution method. This paper studies the more reasonable and effective interpolated convolution operator and solves the problem of uneven point density. Thomas et al.~\cite{9010002} proposed a new kernel point convolution operator on point clouds. It follows the shared MLP based method but uses a set of kernel points in a regular polyhedron disposition to carry the weights. To extract the feature vector of a point, it uses the weighted sum of all the kernel points in its ball neighborhood, and the weights are related with the distances between kernel points and the input point. This is different from our 3D convolution method which focuses on structured 3D convolution.

\par Recently, there have been multiple works combined with more spatial geometric information and more constraints to improve the performance of 3D semantic segmentation~\cite{chiang2019unified,narita2019panopticfusion,choy20194d,jaritz2019multi}. MVPNet~\cite{jaritz2019multi} introduced multi-view image features into 3D point clouds, then used a point clouds based network to predict 3D semantic labels. Joint point-based~\cite{chiang2019unified}, by projecting 2D image features into 3D, proposed a unified fusion optimization framework of pixel-level feature, geometric structure, and global scene context. Then, higher performance was obtained through camera pose information. PanopticFusion~\cite{narita2019panopticfusion} used 2D semantics and instance segmentation information to obtain a certain gain by jointly optimizing 2D image features, 3D structures, and global contexts, and further improved the gain of  \(+1.3\%\)  by synthesizing the pose of the camera. MinkowskiNet~\cite{choy20194d} modeled 4D data with time series from the entire point clouds through camera pose and achieved great performance. These works use more information such as global information or camera poses. However, our method only uses point clouds.

\section{Spherical Interpolated Convolution with Distance-Feature Density}\label{3}

In this section, the specific process of our proposed method is introduced firstly. Then, the spherical interpolated convolution operator for extracting spatial features and the distance-feature density for the non-uniform distribution of point clouds are elaborated respectively. Based on these, the network structure for 3D semantic segmentation of point clouds is proposed.

\subsection{Main Process}\label{section:1}
The feature extraction process of our spherical interpolated convolution kernel is shown in Fig. \ref{fig6}.
\par In the proposed method, each of the basic convolution block works as follows. The positions and features of the ${{N}_{1}}$ input points in each basic convolution block are respectively defined as ${{X}_{1}}=\{{{x}_{i}}\in {{R}^{3}}|i=1,2,...,{{N}_{1}}\}$ and ${{F}_{1}}=\{{{f}_{i}}\in {{R}^{{{c}_{in}}}}|i=1,2,...,{{N}_{1}}\}$. First, the farthest point sampling (FPS) technique is used to sample ${{N}_{2}}$ output points from ${{N}_{1}}$ input points. And according to the position information ${{X}_{1}}$, the position information ${{X}_{2}}=\{{{x}_{j}}\in {{R}^{3}}|j=1,2,...,{{N}_{2}}\}$ corresponding to the ${{N}_{2}}$ points can be obtained. Then each ${{x}_{j}}\in {{X}_{2}}$ is taken as the convolution kernel center, and the positions of the corresponding $K$ convolution cells ${{X}^{(j)}}=\{({{x}_{j}}+\delta {{x}_{k}})\in {{R}^{3}}|k=1,2,...,K\}$ are determined by the spatial structure of the spherical interpolated convolution kernel, where ${\delta {{x}_{k}}}$ is the spatial offset of the ${k}$-th convolution cell center relative to the convolution kernel center ${{x}_{j}}$. The features of the convolution cells ${{F}^{(j)}}=\{{{f}_{k}}^{(j)}\in {{R}^{{{c}_{in}}}}|k=1,2,...,K\}$ are obtained by interpolating the surrounding features in ${{F}_{1}}$ to the cells. The spherical interpolated convolution kernel will be introduced in detail in Section \ref{sec3.2}. At last, the 3D convolution is applied to the spherical interpolated features ${{F}^{(j)}}$ corresponding to each ${{x}_{j}}$. From formula (\ref{(1)}) as follows, the features ${{F}_{2}}=\{{{f}_{j}}\in {{R}^{{{c}_{out}}}}|j=1,2,...,{{N}_{2}}\}$ of ${{N}_{2}}$ output points can be obtained.
\begin{equation}\label{(1)}
	{{f}_{j}}=\sigma(\sum\limits_{k=1}^{K}{w({{x}_{j}}+\delta {{x}_{k}})\cdot {{f}_{k}}^{(j)}}+b_j),
\end{equation}
where $w({{x}_{j}}+\delta {{x}_{k}})$ is the learned weights corresponding to each cell of the convolution kernel. $b_j$ is the learned offset of the convolution kernel. $\sigma$ is the activation function.


\subsection{Hierarchical Spherical Interpolated Convolution Kernel}\label{sec3.2}

In order to deal with the unstructured property of point clouds, InterpCNN~\cite{mao2019interpolated} interpolated the point clouds into corners of a cubeto better extract structured features of point clouds. Considering the asymmetry of cube kernels, a spherical interpolated method with complete spatial symmetry is proposed in this paper.

\par Thanks to the perfect symmetrical structure in 3D space, the spherical interpolated method guarantees that points with the same distance to the regular cell center all contribute to the feature of the cell center. The structure of the spherical interpolated kernel draws on the cubic close-packed model of metal~\cite{hales1998overview}, which proves to have the highest density that can be achieved by any arrangement of spheres~\cite{szpiro2003mathematics,hales2017formal}. However, due to the tangent relationship between each metal atom, which causes gaps between the spheres, directly using the metal cubic close-packed model as the operator kernel will cause some points to be missed. Therefore, each sphere is expanded appropriately in the original position based on the original structure so that they can completely cover the space. This leads to the 4 spherical cells intersect at one centeral point.

\begin{figure}[t]
	\begin{center}
		\includegraphics[width=1.0\linewidth]{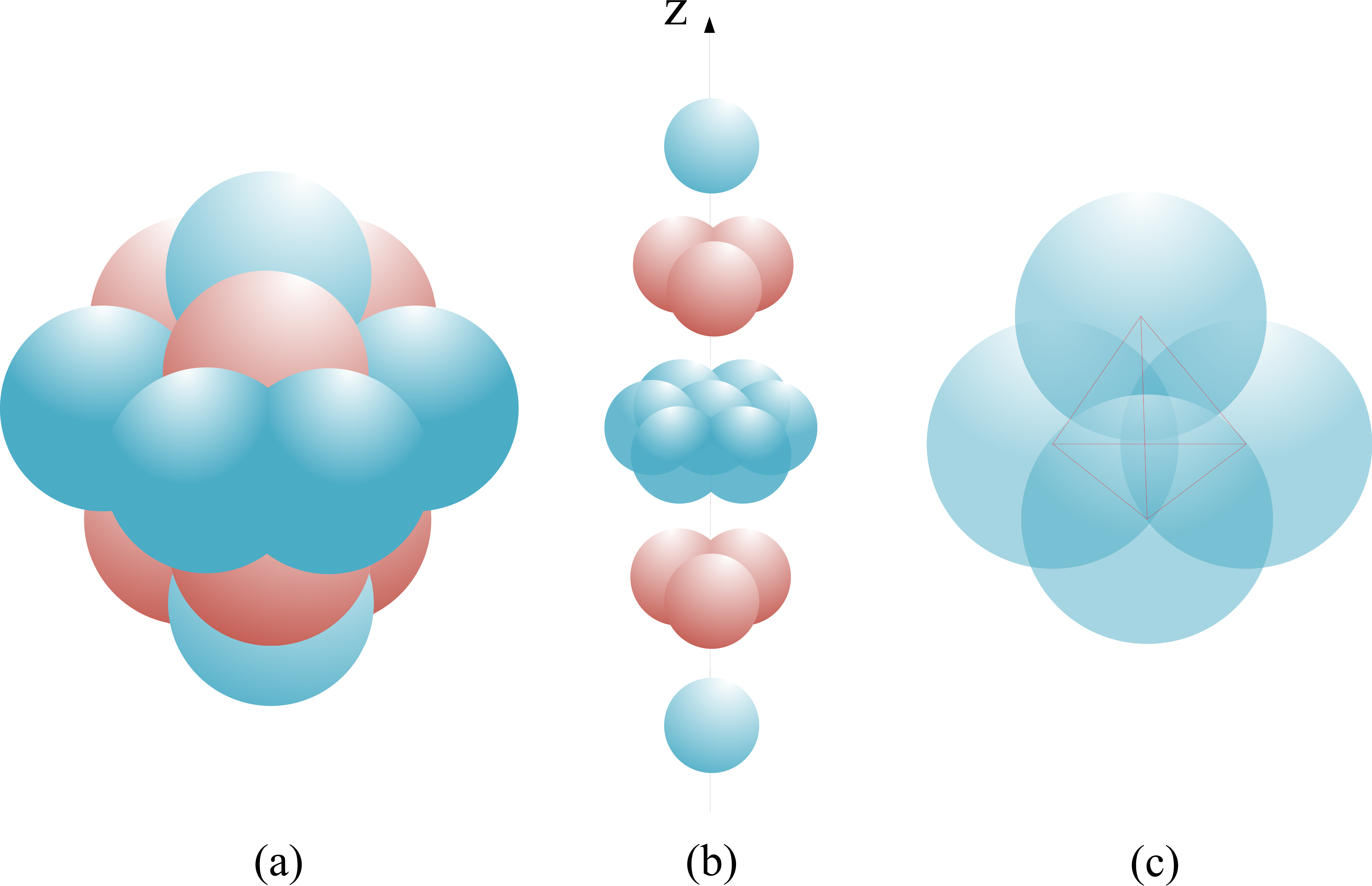}
	\end{center}
	\vspace{-4mm}
	\caption{The schematic diagram of hierarchical spherical interpolated convolution kernel. (a) shows a hierarchical spherical interpolated kernel. (b) shows the hierarchical structure of spherical interpolated kernel. (c) shows a schematic diagram of the relationship between four adjacent cells.}
	\label{fig2}
\end{figure}

As shown in Fig.~\ref{fig2}(a), the complete structure of a spherical interpolated kernel is presented. The spherical interpolated kernel is a closely arranged structure that is vertically symmetrical. In the $z$ direction as shown in Fig.~\ref{fig2}(b), the entire structure can be layered, and the number of spherical cells in each layer is decreased from the middle to the bottom or top. The basic unit of a spherical interpolated kernel in Fig.~\ref{fig2}(c) shows that the centers of four spherical cells constitute a regular tetrahedron, and the four spherical cells intersect at the center of the regular tetrahedron. It can be calculated that the relationship between the radius $r$ of the spherical cell and the side length $l$ of the regular tetrahedron is:
\begin{equation}\label{(2)}
	r=\frac{\sqrt{6}}{4}l.
\end{equation}

\par
\begin{figure}[t]
	\begin{center}
		\includegraphics[width=1.0\linewidth]{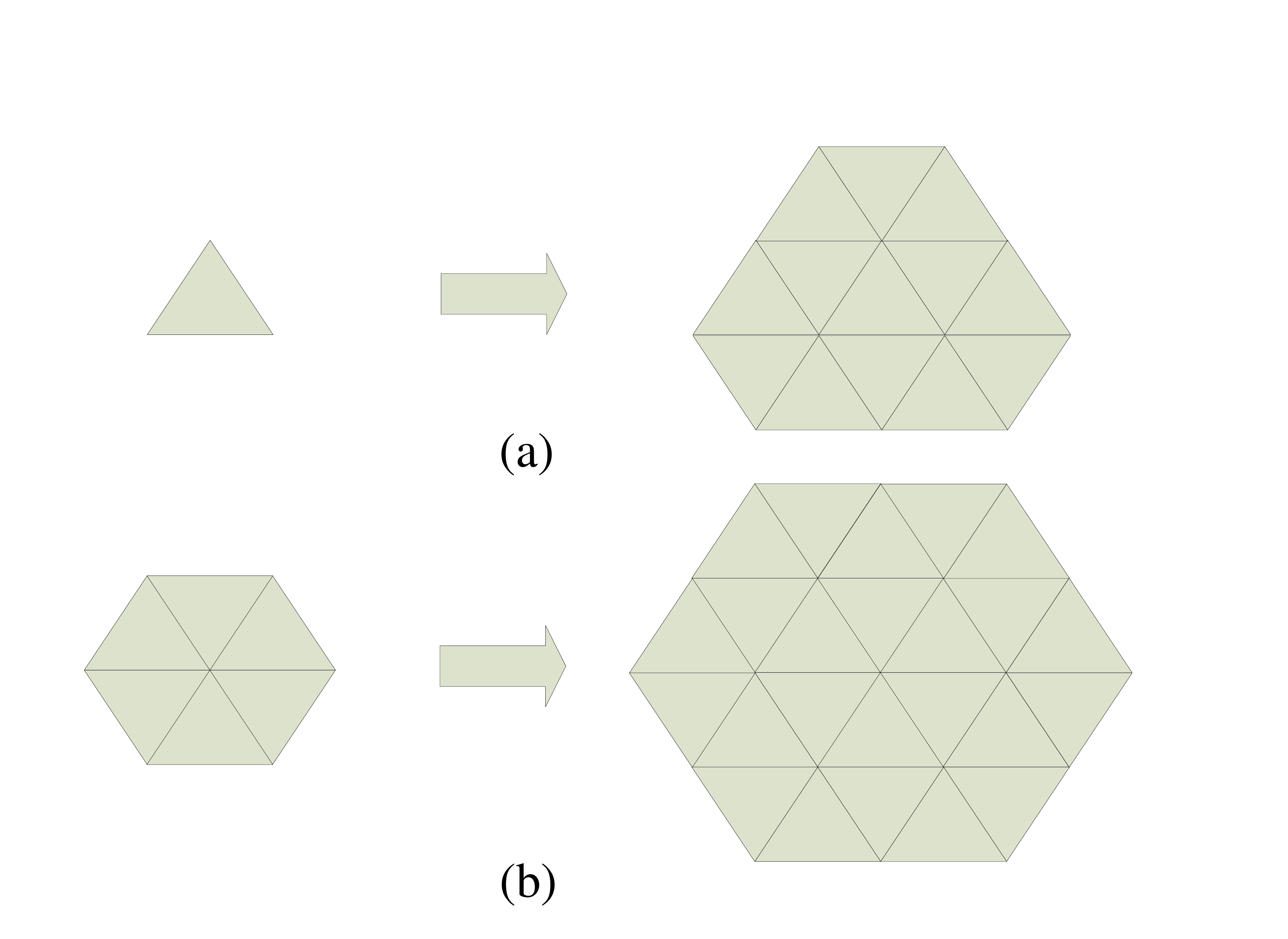}
	\end{center}
	\vspace{-5mm}
	\caption{Two expansion types for the horizontal layer of spherical interpolated convolution kernel.}
	\label{fig3}
\end{figure}

 The position of each spherical cell has a specific calculation method. The height $h$ in $z$ direction between adjacent layers is $\frac{4}{3}r$. The center position of each layer can be calculated by adding or subtracting $h$ on the basis of the adjacent layer. For the cell centers in one layer, the relative positions are determined according to Fig.~\ref{fig3}. The triangles in Fig.~\ref{fig3} are all regular triangles, and side $l$ of each triangle can be calculated by formula (\ref{(2)}): $l=\frac{4}{\sqrt{6}}r$. Each vertex of the triangles represents the position of a convolution cell center. 
\par At the same time, Fig.~\ref{fig3} also shows the horizontal layer expansion perpendicular to the $z$ axis in Fig.~\ref{fig2}(b) in a spherical interpolated kernel. Starting from the sides of the triangles on the outer side of each layer, add the same triangles about the side symmetry; Starting from the vertices of the triangles on the outer side of each layer, add the same triangles about the vertice symmetry. For the expansion of the layer number, when all horizontal layers are expanded, the structure of Fig.~\ref{fig2}(c) is added to the top and the bottom respectively to realize vertical expansion, and each four adjacent spheres still satisfies the structure of Fig.~\ref{fig2}(c).

\par The spherical interpolated kernel is compared with the cube interpolated kernel.

\begin{figure}[t]
	\begin{center}
		\vspace{3mm}
		\includegraphics[width=1.0\linewidth]{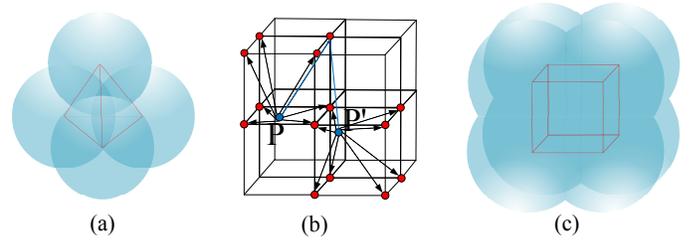}
	\end{center}
	\vspace{-3mm}
	\caption{Comparison of spherical interpolated kernel and cube interpolation kernel. (a) shows the unit of spherical interpolated kernel. (b) shows the InterpConv~\cite{mao2019interpolated} in 3D space. (c) shows the unit of cube interpolation kernel. In the subfigure (b), The $p$ point and $p'$ point have the same distance to the upper center of the cube, but $p$ has the contribution to the upper center and $p'$ does not have the contribution to the upper center.}
	\label{fig4}
\end{figure}

Firstly, the feature on a certain convolution cell is extracted by interpolation, and the feature of the cells is further extracted by 3D convolution to the center of the convolution kernel. The interpConv in InterpCNN~\cite{mao2019interpolated} shown in Fig.~\ref{fig4}(b), due to the spatial geometric nature, cannot completely cover the points of the same distance for each corner. While for the spherical interpolated kernel, points that have the same distance to the regular position all have a contribution.

\begin{figure*}[t]
	\begin{center}
		\vspace{2mm}
		\includegraphics[width=0.95\linewidth]{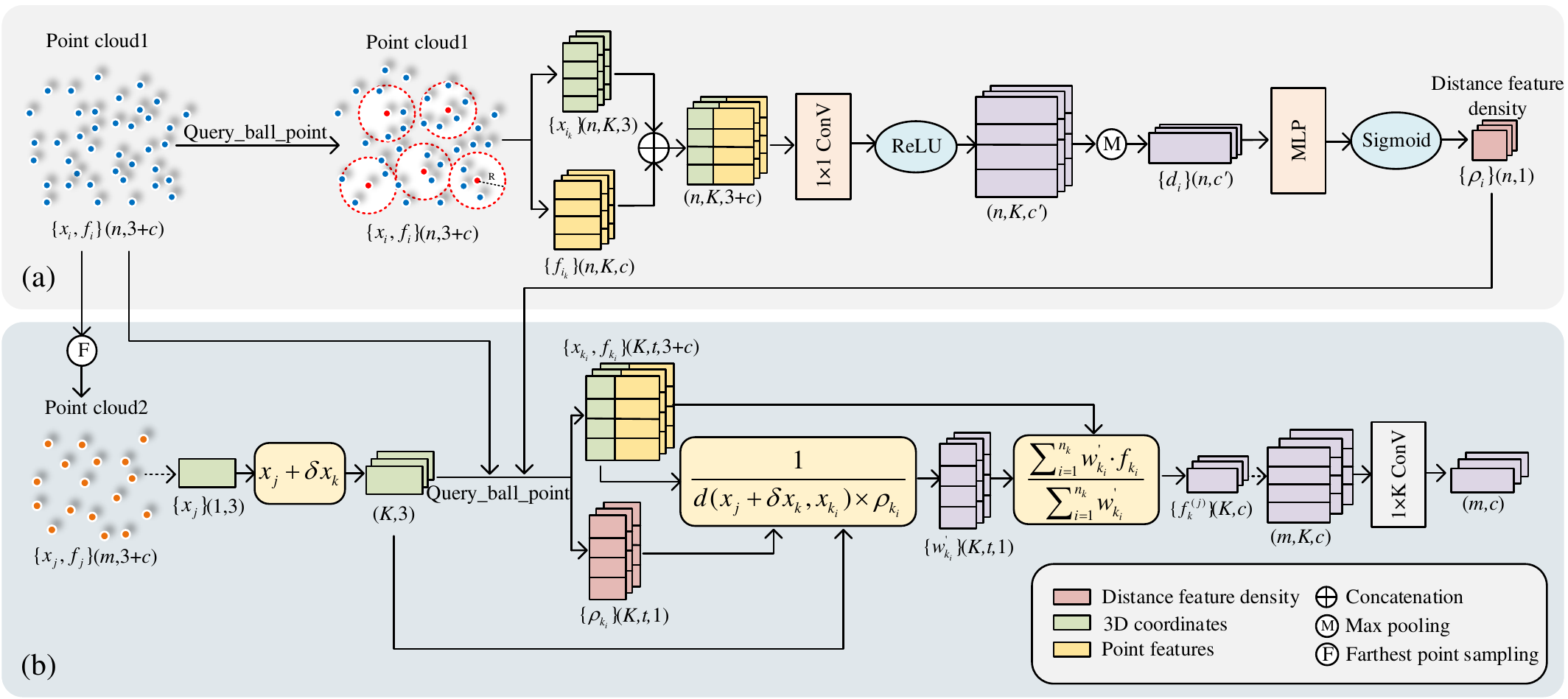}
	\end{center}
	\vspace{0.5mm}
	\caption{Distance-feature density and adjusted spherical interpolated Convolution. (a) Calculating process of distance feature density. (b) Spherical interpolated convolutional operator with distance feature density.}
	\label{ops}
\end{figure*}

\begin{figure}[t]
	\begin{center}
		\includegraphics[width=0.6\linewidth]{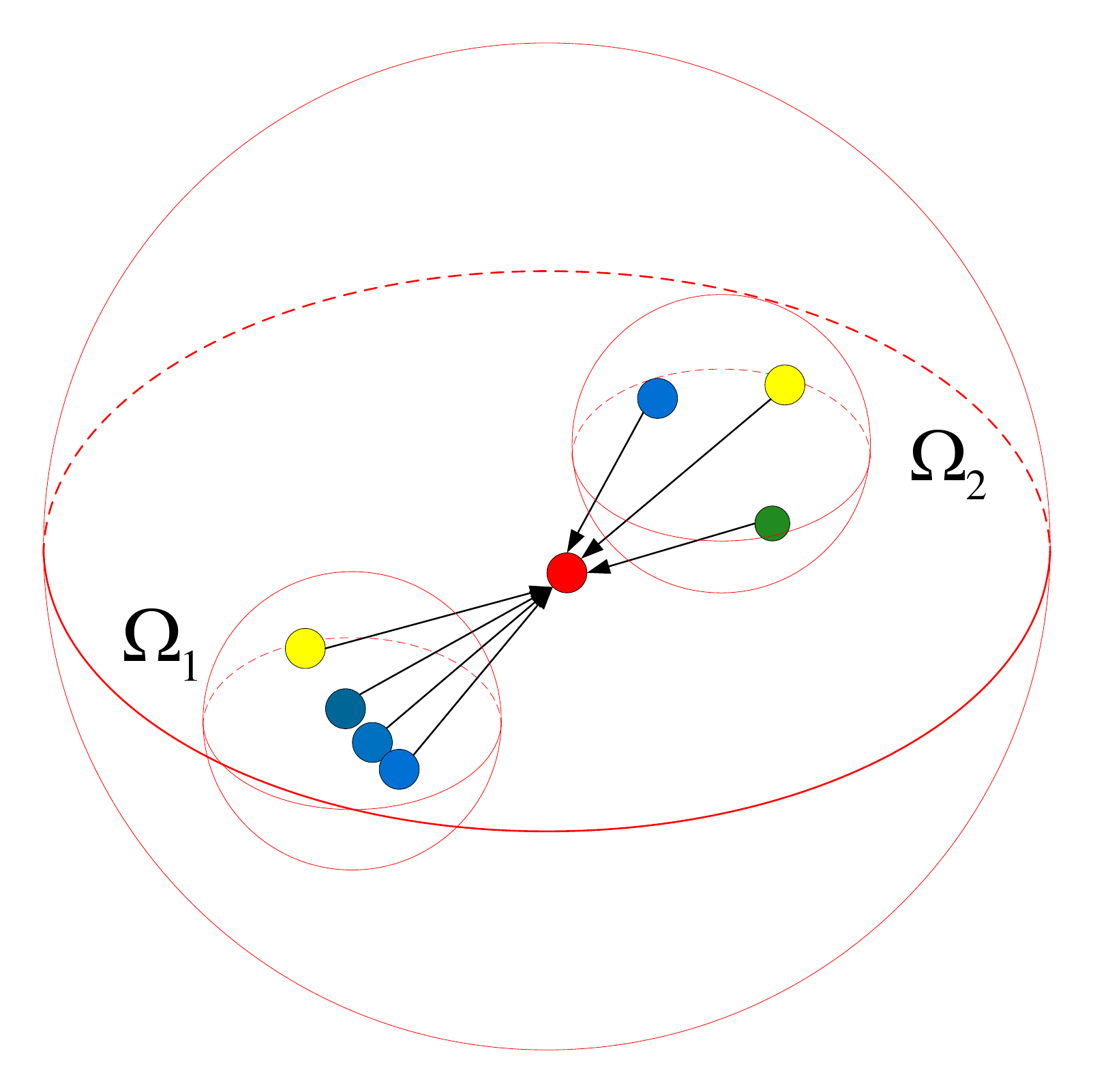}
	\end{center}
	\vspace{-7mm}
	\caption{The schematic diagram for the effect of point cloud feature density on interpolation. ${{\Omega }_{1}}$ is a feature-intensive region, while ${{\Omega }_{2}}$ is a feature-sparse region. Different colors represent different features, and the similarity of colors represents the similarity of features.}
	\label{fig5}
\end{figure}

\par Secondly, dense spherical interpolated kernel has a higher space occupancy than cube interpolated kernel even if the sphere cell is also used in the cube kernel. The basic units of two kernels are shown in Fig.~\ref{fig4}(a) and (c). The radius of spheres is $r$, and the volume of a spherical cap is ${{V}_{sc}}=\pi (r-\frac{h}{3})\cdot {{h}^{2}}$, where ${h}$  is the height of spherical cap. For the basic unit in Fig.~\ref{fig4}(a), $h=(1-\frac{\sqrt{6}}{3})r$. Then the repeating volume between each two spheres is ${{V}_{sp-sc}}={2}{{{V}}_{sc}}=0.199{{r}^{3}}$. The average volume of the space covered by each sphere is ${{V}_{sp}}=(4\times \frac{4}{3}\pi {{r}^{3}}-6\times {{V}_{sp-sc}})/4=3.891{{r}^{3}}$. For the basic unit in Fig.~\ref{fig4}(c), there are two types of the spherical cap. For the spherical cap between the two spheres on the same side of the cube, $h=(1-\frac{\sqrt{3}}{3})r$, and the repeating volume is ${{V}_{sq-sc1}}={2}{{{V}}_{sc}}=0.964{{r}^{3}}$. For the spherical cap between the two spheres on the diagonal of the cube, $h=(1-\frac{\sqrt{6}}{3})r$, and the repeating volume is ${{V}_{sq-sc2}}={2}{{{V}}_{sc}}=0.199{{r}^{3}}$. The average volume of the space covered by each sphere is ${{V}_{sq}}=(8\times \frac{4}{3}\pi {{r}^{3}}-12\times {{V}_{sp-sc1}}-4\times {{V}_{sp-sc2}})/8=2.643{{r}^{3}}$. It can be seen from the calculation results that the repeat volume of the spherical interpolated kernel is smaller than the cube kernel, and the unit sphere of spherical interpolated kernel covers a larger volume. The above comparison shows that for the same size of the space to extract features, the unit sphere of the spherical interpolated kernel covers a larger volume than the cube kernel. Therefore, fewer spherical cells are needed to cover the same space for the spherical interpolated kernel, which reduces the number of convolution kernel weight parameters.

\par In the interpolation process, the traditional KNN and classic ball query algorithms are not used. The operator used is to interpolate all points in a spherical interpolated kernel to the regular positions, that is the cell centers. This approach preserves the integrity of the position information. Linear interpolation~\cite{qi2017pointnet++} is adopted in the interpolation process:
\begin{equation}\label{(3)}
	{{f}_{k}}^{(j)}=\frac{\sum\nolimits_{i=1}^{{{n}_{k}}}{{{w}_{{{k}_{i}}}}\cdot {{f}_{{{k}_{i}}}}}}{\sum\nolimits_{i=1}^{{{n}_{k}}}{{{w}_{{{k}_{i}}}}}},
	{{w}_{{{k}_{i}}}}=\frac{1}{d{{({{x}_{j}}+\delta {{x}_{k}},{{x}_{{{k}_{i}}}})}^{2}}},
\end{equation}
where ${{x}_{j}}\in {{{X}}_{2}}$. ${{x}_{{{k}_{i}}}}\in {{X}_{1}},{{f}_{{{k}_{i}}}}\in {{F}_{1}}$ represent the position and feature information of the points contained in the $k$-th cell among the ${{N}_{1}}$ input points. ${{n}_{k}}$ represents the number of points contained in the $k$-th cell.

\par However, there are still some problems with the interpolation method. As shown in Fig.~\ref{fig5}, in the interpolation process, due to the non-uniform nature of the point clouds, there are some local regions like ${{\Omega }_{1}}$ with dense points. The features of the dense points are similar or even tend to be the same, while other local regions like ${{\Omega }_{2}}$ are sparse and the features are not highly similar. At the same time, in the process of down-sampling point clouds, there is a problem with repetitive sampling of point clouds. For these problems, a method of distance-feature density is proposed in the interpolation process to adjust the interpolation weights which is different from PointConv~\cite{wu2019pointconv}. PointConv only takes the density of points into account. The specific analysis is as described in Section \ref{sec3.3}.


\subsection{Distance-Feature Density and Adjusted Spherical Interpolation}\label{sec3.3}

Considering the repetitive sampling problem for the points that may occur in the down-sampling process and the non-uniform nature of the point clouds, the distance-feature density is proposed to solve these problems. Since the repetitive sampling points also cause the non-uniform nature of point clouds, the method to deal with the non-uniform nature of point clouds is discussed in detail as follows.
\par As shown in Fig.~\ref{fig5}, for region ${{\Omega }_{1}}$, dense point clouds and highly repetitive features make this region have a greater impact on the interpolation, which will result in inaccuracies in the interpolation. 

\par Distance-feature density is proposed to extract the feature of the regular position better. The distance density and feature density of each point are determined by the points in a small local region. To calculate the distance-feature density of $x_i$ ($x_i\in {{X}_{1}}$), ball query \cite{qi2017pointnet++} is used to collect distance information $X_{i}$ and feature information $F_{i}$ of $K$ neighboring points in a local sphere, where ${{X}_{i}}=\{{{x}_{i_k}}\in {{R}^{3}}|k=1,2,...,K\}$ and ${{F}_{i}}=\{{{f}_{i_k}}\in {{R}^{{{c}_{in}}}}|k=1,2,...,K\}$. $k$ indicates the $k$-th neighboring point and $K$ indicates the fixed number of queried points. Then $1 \times 1$ convolution and ReLU are adopted to discover the inner relationship between ${{X}_{i}}$ and ${{F}_{i}}$. After that, max pooling is adopted to extract the aggregated density feature $d_{i}$ for $x_i$. Then, density feature $d_{i}$ is sent to an MLP, and the distance-feature density ${{\rho _{i}}}$ is obtained as the output of sigmoid activation function. The overall calculating process for distance-feature density is illustrated in Fig.~\ref{ops}(a).

\begin{equation}\label{mlp}
	\begin{split}
		d_{i}&=maxpooling(Relu(Conv({{X}_{i}} \oplus {{F}_{i}}))),\\
		{{\rho _{i}}}&=Sigmoid(MLP(d_{i})).
	\end{split}	
\end{equation}

\par When ${{f}_{{{k}_{i}}}}$ in formula (\ref{(3)}) is interpolated to ${{f}_{k}}^{(j)}$, in order to consider the influence of the non-uniform nature of the point clouds, the weight of the interpolation is multiplied by the inverse of the distance-feature density ${{\rho _{{k}_{i}}}}$ to tune the contribution of ${{f}_{{{k}_{i}}}}$ to ${{f}_{k}}^{(j)}$. That is, the formula (\ref{(3)}) can be updated to (\ref{(4)}), and the modified spherical interpolated convolution process with distance-feature density is shown in Fig.~\ref{ops}(b).
\begin{equation}\label{(4)}
	\begin{split}
		&{{f}_{k}}^{(j)}=\frac{\sum\nolimits_{i=1}^{{{n}_{k}}}{{{w}_{{{k}_{i}}}'}\cdot {{f}_{{{k}_{i}}}}}}{\sum\nolimits_{i=1}^{{{n}_{k}}}{{{w}_{{{k}_{i}}}'}}},\\
		&{{w}_{{{k}_{i}}}'}=\frac{1}{d{{({{x}_{j}}+\delta {{x}_{k}},{{x}_{{{k}_{i}}}})}^{2}}\cdot {{\rho _{{k}_{i}}}}}.
	\end{split}
\end{equation}

\begin{figure*}[t]
	\begin{center}
		\vspace{0mm}
		\includegraphics[width=0.98\linewidth]{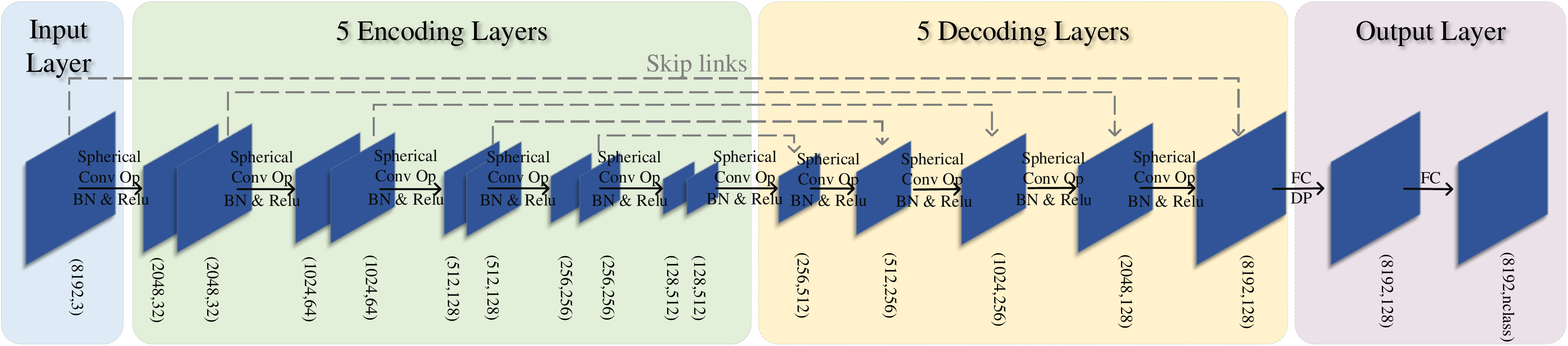}
	\end{center}
	\vspace{-2mm}
	\caption{Illustration of our spherical interpolated convolutional network for semantic segmentation of 3D point clouds. FC means Fully Connected Layer. DP means Dropout. BN means Batch Normalization. Spherical Conv Op meas our proposed spherical interpolated convolution operator.}
	\label{na}
\end{figure*}

\par For similar or even the same input points in a local region like in Fig.~\ref{fig5}, the adjust method is explained according to a special case as follows. Assume that two points in ${{N}_{1}}$ points is the same points, and their position and feature information are ${{x}_{{{k}_{1}}}},{{x}_{{{k}_{2}}}},{{f}_{{{k}_{1}}}},{{f}_{{{k}_{2}}}}$. From the assumption, it can be known that ${{x}_{{{k}_{1}}}}={{x}_{{{k}_{2}}}},{{f}_{{{k}_{1}}}}={{f}_{{{k}_{2}}}},{{w}_{{{k}_{1}}}}={{w}_{{{k}_{2}}}}$. For the ideal situation, these two points are only needed to be calculated once, that is:
\begin{equation}\label{(5)}
	\begin{split}
		{{f}_{{{k}_{id}}}}^{(j)}=&\frac{{{w}_{{{k}_{1}}}}\cdot {{f}_{{{k}_{1}}}}}{{{w}_{{{k}_{1}}}}+\sum\nolimits_{i=3}^{{{n}_{k}}}{{{w}_{{{k}_{i}}}}}}+\frac{\sum\nolimits_{i=3}^{{{n}_{k}}}{{{w}_{{{k}_{i}}}}\cdot {{f}_{{{k}_{i}}}}}}{{{w}_{{{k}_{1}}}}+\sum\nolimits_{i=3}^{{{n}_{k}}}{{{w}_{{{k}_{i}}}}}},
	\end{split}
\end{equation}
while the actual feature before tuning the weights is:
\begin{equation}\label{(6)}
	\begin{split}
		{{f}_{{{k}_{ac}}}}^{(j)}=\frac{{{w}_{{{k}_{1}}}}{{f}_{{{k}_{1}}}}+{{w}_{{{k}_{2}}}}{{f}_{{{k}_{2}}}}}{\sum\nolimits_{i=1}^{{{n}_{k}}}{{{w}_{{{k}_{i}}}}}}+\frac{\sum\nolimits_{i=3}^{{{n}_{k}}}{{{w}_{{{k}_{i}}}}{{f}_{{{k}_{i}}}}}}{\sum\nolimits_{i=1}^{{{n}_{k}}}{{{w}_{{{k}_{i}}}}}}.
	\end{split}
\end{equation}
\par In order to make the actual calculated ${{f}_{{{k}_{ac}}}}^{(j)}$ equal to the ideal ${{f}_{{{k}_{id}}}}^{(j)}$, from the formulas (\ref{(5)}) and (\ref{(6)}), it is only needed to multiply ${{w}_{{{k}_{1}}}}$ and ${{w}_{{{k}_{2}}}}$ by the coefficient $\frac{1}{2}$ to tune them. Then the tuned ${{f}_{{{k}_{ac}}}}^{(j)}$ is:
\begin{small}
	\begin{equation}\label{(7)}
		\begin{split}
			{{f}_{{{k}_{ac}}}}^{(j)}\!=&\frac{\frac{1}{2}{{w}_{{{k}_{1}}}}{{f}_{{{k}_{1}}}}\!+\!\frac{1}{2}{{w}_{{{k}_{2}}}}{{f}_{{{k}_{2}}}}}{\frac{1}{2}{{w}_{{{k}_{1}}}}\!+\!\frac{1}{2}{{w}_{{{k}_{2}}}}\!+\!\sum\nolimits_{i=3}^{{{n}_{k}}}{{{w}_{{{k}_{i}}}}}}\!+\!\frac{\sum\nolimits_{i=3}^{{{n}_{k}}}{{{w}_{{{k}_{i}}}}{{f}_{{{k}_{i}}}}}}{\frac{1}{2}{{w}_{{{k}_{1}}}}\!+\!\frac{1}{2}{{w}_{{{k}_{2}}}}\!+\!\sum\nolimits_{i=3}^{{{n}_{k}}}{{{w}_{{{k}_{i}}}}}}\\
			=&{{f}_{{{k}_{id}}}}^{(j)}.
		\end{split}
	\end{equation}
\end{small}
So that, the tuned ${{w}_{{{k}_{1}}}}$ and ${{w}_{{{k}_{2}}}}$ are:
\begin{equation}\label{(8)}
	{{w}_{{{k}_{1}}}'}={{w}_{{{k}_{2}}}'}=\frac{1}{2}{{w}_{{{k}_{1}}}}=\frac{1}{2}{{w}_{{{k}_{2}}}}.
\end{equation}
Therefore, the distance-feature densities are: 
\begin{equation}\label{(9)}
	{{\rho _{d - f}}}({{x}_{{{k}_{1}}}})={{\rho _{d - f}}}({{x}_{{{k}_{2}}}})=2.
\end{equation}

\par From the special case, it is shown that with inverse distance-feature density multiplied by ${{w}_{{{k}_{i}}}}$, the calculated feature in the regular position can be closer to the actual feature.

\subsection{Spherical Interpolated Convolutional Network}\label{subsec3.4}
 The spherical interpolated convolutional network is designed combined with the spherical interpolated convolution kernel for the semantic segmentation task. Following the typical encoder-decoder structure of U-Net~\cite{ronneberger2015u}, the proposed network consists of a 5-layer encoding part and a 5-layer decoding part and uses fully connected (FC) layers with dropout to predict the final result, as shown in Fig.~\ref{na}.

\subsubsection{5 Encoding Layers}
Each layer of the encoding part consists of two spherical interpolated convolution operators. The first spherical interpolated convolution operator of each layer is to downsample ${{N}_{1}}$ input points to ${{N}_{2}}$ output points, and extract features from ${{F}_{1}}$ to ${{F}_{2}}$. Then ${{F}_{2}}$ of ${{N}_{2}}$ output points are taken as input to the next operator. In the second operator, the downsampling is not performed, but the feature is further extracted by the spherical interpolated convolution operator. The shape transition in one layer is (${{N}_{1}}$,${{F}_{1}}$)$\rightarrow$(${{N}_{2}}$,${{F}_{2}}$)$\rightarrow$(${{N}_{2}}$,${{F}_{2}'}$). The batch normalization and ReLU activation are used after each convolution layer. 

\subsubsection{5 Decoding Layers}
The decoding part is also a 5-layer convolution structure. Different from the encoding layers, each decoding layer only uses a spherical interpolated convolution operator. It is used at first to extract the feature ${{F}_{1}'}$ of ${{N}_{1}}$ output points from the ${{N}_{2}}$ input feature ${{F}_{2}'}$. Then the skip link is used to pass the feature ${{F}_{1}}$ to concatenate with ${{F}_{1}'}$. After the concatenation, a MLPs are used to further extract feature ${{F}_{1}''}$ of the ${{N}_{1}}$ output points. The shape transition in one decoding layer is (${{N}_{2}}$,${{F}_{2}'}$) $\rightarrow$ (${{N}_{1}}$,${{F}_{1}'}$) $\rightarrow$ (${{N}_{1}}$,${{F}_{1}'}+{{F}_{1}}$) $\rightarrow$ (${{N}_{1}}$,${{F}_{1}''}$). The batch normalization and ReLU activation are also used here. 

\subsubsection{Output Layers}
Two FC layers are used after the decoding part to predict the semantic information of each point, and the first with a 0.5 probability dropout. The output size of the network is $(N,nclass)$. The number of points $N$ is 8192, which is the same as the input.


\setlength{\tabcolsep}{0.8mm}{
	\begin{table*}[t]
		\centering
		\vspace{2mm}
		\caption{Semantic segmentation results on ScanNet test set~\cite{dai2017scannet}. Mean IoU over
			categories (mIOU) is reported.}
		\begin{tabular}{c|c|cccccccccccccccccccc}
			\hline
			Method& mIOU & wall & floor & cab  & bed  & chair & sofa & table & door & wind & bkshf & pic  & cntr & desk & curt & fridg & showr & toil & sink & bath & ofurn \\ \hline
			ScanNet~\cite{dai2017scannet}               & 30.6 & 43.7 & 78.6  & 31.1 & 36.6 & 52.4  & 34.8 & 30.0  & 18.9 & 18.2 & 50.1  & 10.2 & 21.1 & 34.2 & 0.2  & 24.5  & 15.2  & 46.0 & 31.8 & 20.3 & 14.5  \\
			PointNet++~\cite{qi2017pointnet++}          & 33.9 & 52.3 & 67.7  & 25.6 & 47.8 & 36.0  & 34.6 & 23.2  & 26.1 & 25.2 & 45.8  & 11.7 & 25.0 & 27.8 & 24.7 & 21.2  & 14.5  & 54.8 & 36.4 & 58.4 & 18.3  \\
			SPLATNet~\cite{su2018splatnet}            & 39.3 & 69.9 & 92.7  & 31.1 & 51.1 & 65.6  & 51.0 & 38.3  & 19.7 & 26.7 & 60.6  & 0.0  & 24.5 & 32.8 & 24.5 & 0.1   & 24.9  & 59.3 & 27.1 & 47.2 & 22.7  \\
			Tangent Convolutions~\cite{tatarchenko2018tangent} & 43.8 & 63.3 & 91.8  & 36.9 & 64.6 & 64.5  & 56.2 & 42.7  & 27.9 & 35.2 & 47.4  & 14.7 & 35.3 & 28.2 & 25.8 & 28.3  & 29.4  & 61.9 & 48.7 & 43.7 & 29.8  \\
			3DMV~\cite{dai20183dmv}             & 48.4 & 60.2 & 79.6  & 42.4 & 53.8 & 60.6  & 50.7 & 41.3  & 37.8 & 53.9 & 64.3  & 21.4 & 31.0 & 43.3 & 57.4 & 53.7  & 20.8  & 69.3 & 47.2 & 48.4 & 30.1  \\
			PanopticFusion~\cite{narita2019panopticfusion}      & 52.9 & 60.2 & 81.5  & 38.6 & 68.8 & 63.2  & 64.9 & 44.2  & 29.3 & 56.1 & 60.4  & 24.1 & 22.5 & 43.4 & 70.5 & 49.9  & 66.9  & 79.6 & 50.7 & 49.1 & 34.8  \\
			PointConv~\cite{wu2019pointconv}         & 55.6 & 76.2 & 94.4  & 47.2 & 64.0 & 73.9  & 63.9 & 50.5  & 44.5 & 51.5 & 57.4  & 18.5 & 43.0 & 41.8 & 43.3 & 46.4  & 57.5  & 82.7 & 54.0 & 63.6 & 37.2  \\
			TextureNet~\cite{huang2019texturenet}          & 56.6 & 68.0 & 93.5  & 49.4 & 66.4 & 71.9  & 63.6 & 46.4  & 39.6 & 56.8 & 67.1  & 22.5 & 44.5 & 41.1 & 67.8 & 41.2  & 53.5  & 79.4 & 56.5 & 67.2 & 35.6  \\
			SPH3D-GCN~\cite{lei2019spherical}            & 61.0 & 77.3 & 93.5  & 53.2 & 77.2 & 79.2  & 70.5 & 54.9  & 50.7 & 53.4 & 48.9  & 4.6  & 40.4 & 57.0 & 64.3 & 51.0  & 70.2  & 85.9 & 60.2 & 85.8 & 41.4  \\
			HPEIN~\cite{jiang2019hierarchical}             & 61.8 & 80.6 & 94.6  & 59.7 & 66.8 & 76.6  & 61.7 & 57.0  & 52.5 & 60.5 & 64.7  & 21.5 & 41.4 & 52.0 & 68.0 & 49.3  & 59.9  & 89.7 & 63.8 & 72.9 & 43.2  \\
			MCCNN~\cite{hermosilla2018monte}          & 63.3 & 79.5 & 94.9  & 57.6 & 73.1 & 80.9  & 68.0 & 54.2  & 49.1 & 61.8 & 77.1  & 10.5 & 41.0 & 49.7 & 68.4 &58.1  & 64.6  & 81.7 & 62.0 & 86.6 & 46.6  \\
			joint point-based~\cite{chiang2019unified}    & 63.4 & 81.4 & 95.1  & 63.3 & 77.8 & 82.5  & 76.4 & 55.9  & 56.1 & 59.8 & 66.7  & 29.1 & 42.0 & 46.7 & 80.4 & 56.6  & 45.8  & 83.8 & 57.9 & 61.4 & 49.4  
			\\ \hline
			Ours                 & 63.6 & 79.9 & 95.1  & 57.2 & 69.7 & 78.0  & 68.2 & 54.1  & 53.0 & 59.4 & 75.2  & 17.0 & 44.5 & 52.9 & 71.6 & 50.7  & 66.6  & 88.6 & 63.6 & 83.0 & 44.6  \\ \hline
		\end{tabular}
	\end{table*}
}

\section{Experiments and Evaluation}\label{4}

The realistic datasets are used in our experiments, which contains a lot of noise. At the same time, to better prove the robustness of the proposed network, the method is tested on an indoor ScanNet dataset~\cite{dai2017scannet} and an outdoor Paris-Lille-3D dataset~\cite{roynard2018paris} for the semantic segmentation task.

\subsection{Implementation}
Following PointConv~\cite{wu2019pointconv}, a fixed-size cube is used to collect points in every scene during training and validation process. 8192 points are sampled from the cube as the input to the network. When validating or testing the whole scene, the whole scene is scanned and evaluated by a sliding window and a certain overlap is set to make sure that all points can be fetched and given a predicted value.

In the calculation of distance-feature density (in formula \eqref{mlp}), $i$-th $(i = 1, 2,..., 5)$ layer of the encoding part has a fixed query ball radius $R_i$, which is the same as the radius of spherical cells in this layer. It is 0.045 and increases layer by layer in the encoding part. The radius of adjacent layers satisfies this relation: $R_{i+1}=2 \times R_i$. The decoding layer which has the same point number with the encoding layer has the same radius. Besides, the point always queries 64 points within its ball-shaped neighborhood when calculating distance-feature densities. The output channel of the $1 \times 1$ convolution used in formula \eqref{mlp} is fixed at 32. The MLP to gain the distance-feature density with the output channel 16 and 1 before sigmoid activation.

The experiments are all performed on a RTX 2080Ti GPU. The Adam optimizer is used. The batch size is 16. The dimension of each layer is illustrated in Fig.~\ref{na}, and the probability of the dropout is 0.5.

\setlength{\tabcolsep}{2.6mm}{
	\begin{table*}[t]
		\centering
		\vspace{2mm}
		\caption{Semantic segmentation results on Paris-Lille-3D test set~\cite{roynard2018paris}. Mean IoU over categories (mIOU) is reported.}
		\begin{tabular}{c|c|ccccccccc}
			\hline
			Method& mIOU & ground & building & pole & bollard & trash can & barrier & pedestrian & car  & natural \\ \hline
			RF MSSF~\cite{thomas2018semantic}  & 56.3 & 99.3   & 88.6     & 47.8 & 67.3    & 2.3       & 27.1    & 20.6       & 74.8 & 78.8    \\
			MS3 DVS~\cite{roynard2018paris}  & 66.9 & 99.0   & 94.8     & 52.4 & 38.1    & 36.0      & 49.3    & 52.6       & 91.3 & 88.6    \\
			HDGCN~\cite{liang2019hierarchical}    & 68.3 & 99.4   & 93.0     & 67.7 & 75.7    & 25.7      & 44.7    & 37.1       & 81.9 & 89.6    \\ \hline
			Ours     & 70.1 & 99.4   & 95.0     & 65.4 & 71.5    & 22.8      & 43.2    & 48.9       & 95.2 & 89.3    \\ \hline
		\end{tabular}
	\end{table*}
}

\begin{figure}[t]
	\centering
	
	
	\subfigure{\includegraphics[width=0.45\linewidth]{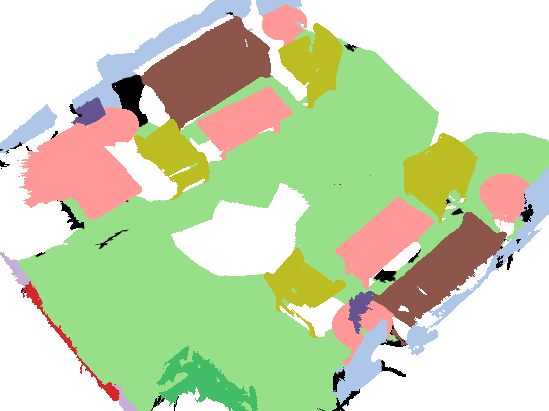}}
	\subfigure{\includegraphics[width=0.45\linewidth]{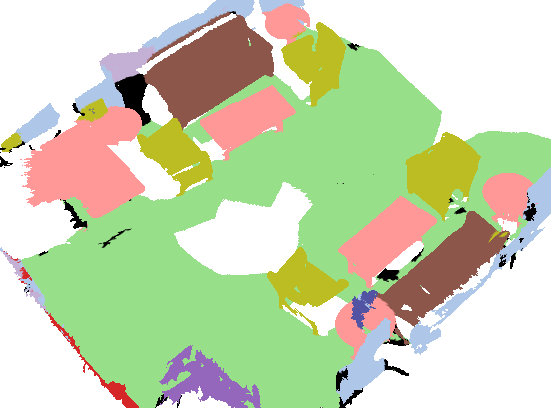}}
	
	\subfigure{\includegraphics[width=0.45\linewidth]{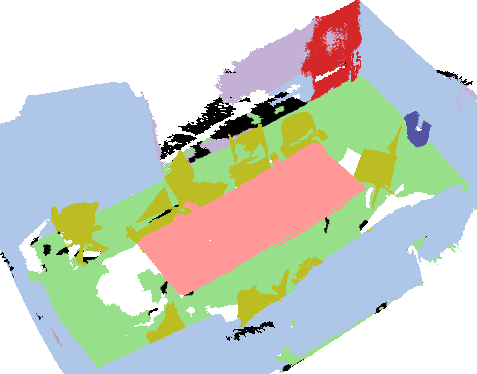}}
	\subfigure{\includegraphics[width=0.45\linewidth]{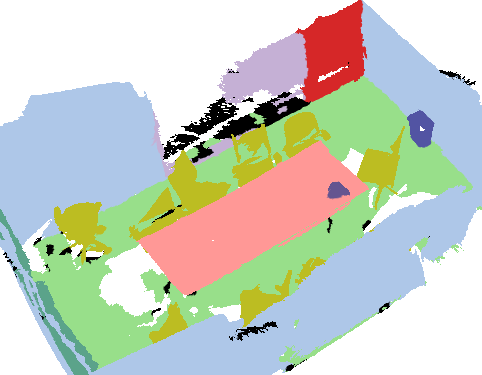}}
	
	\caption{Visualized results on ScanNet dataset~\cite{dai2017scannet}. The ground truth is on the left and prediction is on the right.}
	\label{fig:visu1}
\end{figure}

\subsection{Indoor ScanNet Dataset}\label{scannet}

ScanNet benchmark dataset~\cite{dai2017scannet} uses RGB-D cameras for indoor reconstruction. The dataset has 1513 scenes with true values and 100 testing scenes without true values. Test results for test scenes are given by submitting
the predicted results to the evaluation server. The 1513 training scenes are divided into 1201 training scenes and 312 validation scenes. The size of the cube is $3m\times 1.5m\times 1.5m$, and the overlap is $0.5m$. The results are shown in Table I.  SPH3D-GCN~\cite{lei2019spherical}, HPEIN~\cite{jiang2019hierarchical} are the recent methods combining with graph convolution. Joint point-based~\cite{chiang2019unified} combined global scene context, 2D image features and 3D features for point cloud semantic segmentation. The proposed method surpasses them all.

Joint point-based~\cite{chiang2019unified} needs to establish global maps, while the proposed method is for practical application and does not need extra processing for point clouds or extra information. The proposed method is more robust to the uneven distribution of point clouds, which is very common in the actual application of robots. Some visualization results are shown in Fig. \ref{fig:visu1}.

\subsection{Outdoor Paris-Lille-3D Dataset}
\begin{figure}[t]
	\centering
	
	\subfigure{\includegraphics[width=0.45\linewidth]{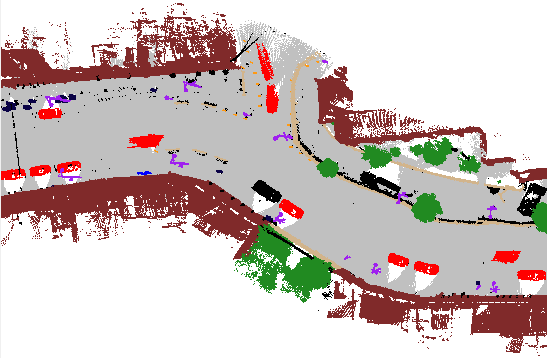}}
	\subfigure{\includegraphics[width=0.45\linewidth]{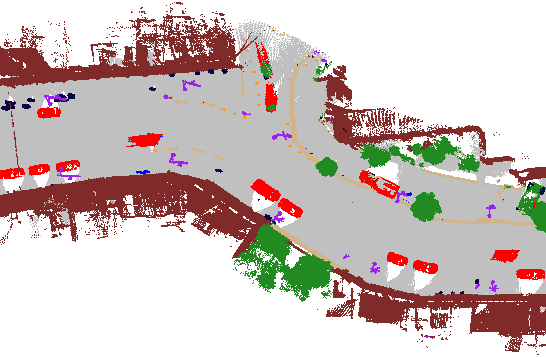}}
	
	\subfigure{\includegraphics[width=0.45\linewidth]{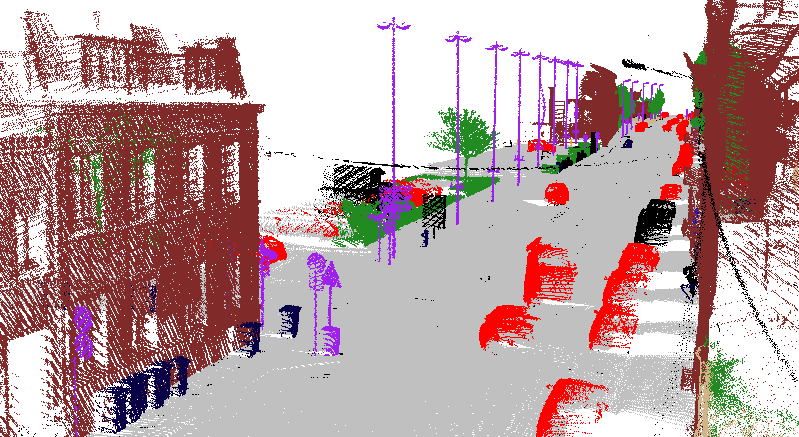}}
	\subfigure{\includegraphics[width=0.45\linewidth]{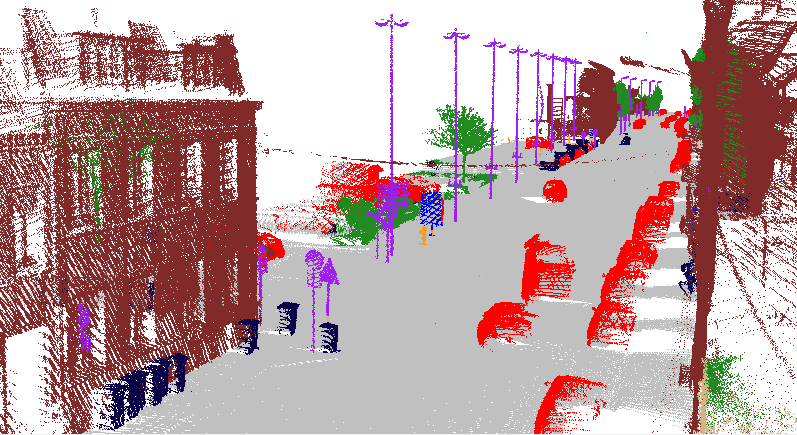}}
	

	\caption{Visualized results on Paris-Lille-3D dataset~\cite{roynard2018paris}. The ground truth is on the left and prediction is on the right.}
	\label{fig:visu}
\end{figure}

Outdoor Paris-Lille-3D dataset~\cite{roynard2018paris} is a large point cloud dataset generated from a Mobile Laser System (MLS) in Paris
and Lille. This dataset includes 4 training scenes and 3 test scenes.
In the experiment, one scene (Lille1\underline{~~}2) is selected from 4
training scenes for validation, and the rest three are used for
training. Unlike the ScanNet dataset, Paris-Lille-3D has only
coordinates as inputs without RGB. The
training scenes are first divided  into small scenes with a length of $10m$ so that a batch of training data is from different scenes like ScanNet. Since point clouds collected by laser in this
dataset are sparser than ScanNet based on RGB-D, the cubes used for data acquisition are adjusted to $10m\times 5m \times 5m$, and overlap is $2.5m$. The results are shown in Table II. It can be seen that the
proposed method exceeds the recent method based on graph network~\cite{liang2019hierarchical}.
The proposed network and HDGCN both use hierarchical structure, which
demonstrates the proposed spherical interpolated convolution kernel is more effective than the DGConv block~\cite{liang2019hierarchical}. Some visualization results are shown in Fig. \ref{fig:visu}.

\subsection{Ablation Study on ScanNet}
We remove or change the key components of our method to do ablation studies and verify the effectiveness of our method. These experiments are tested on the ScanNet validation set, and the implementation is the same as in  Section~\ref{scannet}. The results are shown in Table III.
Experiments show that the proposed distance-feature density and spherical interpolated convolutional kernel both improve the experimental results. The proposed spherical interpolated convolution achieves a slightly better effect than the cube kernel, but the spherical interpolated kernel has 15 trainable kernel weights, and the number of kernel weights in cube kernel is 27. The proposed method reduced the parameters and achieved better performance.

\begin{table}[t]
	\begin{center}
		\caption{Results of ablation studies on ScanNet validation set~\cite{dai2017scannet}. Mean IoU over categories (mIOU) is reported. Cubic means using cubic convolution kernel. Sphere means using spherical interpolated convolution kernel. DFD means distance-feature density. RGB means using RGB information.}
		\begin{tabular}{l|c}
			\hline
			Method& mIOU \\ \hline
			Basic + Cubic           & 60.5     \\
			Basic + Sphere           & 60.9     \\
			Basic + Sphere + DFD     & 61.3     \\
			Basic + Sphere + DFD + RGB & 61.1     \\ \hline
		\end{tabular}
	\end{center}
	\label{5}
\end{table}

\begin{table}[t]
	\begin{center}
		\caption{The parameter quantity and speed of the sphere kernel and cubic kernel.}
		\begin{tabular}{c|ccc}
			\hline
			Kernel & Parameters & FLOPs & Runtime/8192 points \\ \hline
			Cubic &   15.34M    &  7.43G &       397ms          \\
			Sphere &   9.06M     &  5.10G &      264ms           \\ \hline
		\end{tabular}
	\end{center}
	\label{5}
\end{table}

The parameter quantity and speed of the sphere kernel and cubic kernel are as shown in Table IV. It can be seen that, consistent with the theoretical analysis, the spherical interpolated convolution kernel is smaller than the cubic kernel. The spherical interpolated kernel is more memory-efficient and time-efficient.

\subsection{Discussion}
The experiment results have shown that the proposed method outperforms the recent works~\cite{narita2019panopticfusion,wu2019pointconv,huang2019texturenet,lei2019spherical,jiang2019hierarchical,chiang2019unified} in ScanNet dataset and the recent works~\cite{thomas2018semantic,roynard2018paris,liang2019hierarchical} in Paris-Lille-3D dataset. Since the proposed method extracts the local features in a structured way, the method can retain more local spatial distribution and structure information. In addition, the robustness of the proposed method is ensured by the distance-feature density. Thus, the proposed method is more robust to uneven point distribution and local sparsity problems. Since points obtained by RGB-D cameras indoor are only distributed on the surface of the objects and the points obtained by LiDAR outdoor are inherently uneven due to the different object distances and unneglected noises, the robustness to uneven point distributions is of great importance and contributes to ensuring the practical performance in real applications.

\section{Conclusion}
Inspired by the cubic close-packed model of metal~\cite{hales1998overview,szpiro2003mathematics,hales2017formal}, a spherical interpolated convolution kernel is proposed for point clouds. It is proved by analysis and calculation that the proposed kernel has a larger space occupancy than the cube kernel with the same number of trainable weights, which makes feature extraction weights more effectively utilized in 3D space. Experiments show that the novel feature extraction kernel has a better performance than the cube kernel and saves memory. By analyzing the spatial feature interpolation, the distance-feature density is proposed for interpolation, which improves the rationality of point cloud feature extraction and achieves better performance on semantic segmentation. The proposed feature extraction method and distance-feature density are expected to promote practical applications such as unmanned driving and robot navigation.

\ifCLASSOPTIONcaptionsoff
  \newpage
\fi




\bibliographystyle{IEEEtran}  
\bibliography{IEEEabrv,bare_jrnl} 

\begin{thebibliography}{10}
\providecommand{\url}[1]{#1}
\csname url@samestyle\endcsname
\providecommand{\newblock}{\relax}
\providecommand{\bibinfo}[2]{#2}
\providecommand{\BIBentrySTDinterwordspacing}{\spaceskip=0pt\relax}
\providecommand{\BIBentryALTinterwordstretchfactor}{4}
\providecommand{\BIBentryALTinterwordspacing}{\spaceskip=\fontdimen2\font plus
\BIBentryALTinterwordstretchfactor\fontdimen3\font minus
  \fontdimen4\font\relax}
\providecommand{\BIBforeignlanguage}[2]{{%
\expandafter\ifx\csname l@#1\endcsname\relax
\typeout{** WARNING: IEEEtran.bst: No hyphenation pattern has been}%
\typeout{** loaded for the language `#1'. Using the pattern for}%
\typeout{** the default language instead.}%
\else
\language=\csname l@#1\endcsname
\fi
#2}}
\providecommand{\BIBdecl}{\relax}
\BIBdecl

\bibitem{mei2019semantic}
J.~Mei, B.~Gao, D.~Xu, W.~Yao, X.~Zhao, and H.~Zhao, ``Semantic segmentation of
  3d lidar data in dynamic scene using semi-supervised learning,'' \emph{IEEE
  Trans. Intell. Transp. Syst.}, 2019.

\bibitem{xiang2019novel}
B.~Xiang, J.~Tu, J.~Yao, and L.~Li, ``A novel octree-based 3-d fully
  convolutional neural network for point cloud classification in road
  environment,'' \emph{IEEE Trans. Geosci. Remote Sens.}, vol.~57, no.~10, pp.
  7799--7818, 2019.

\bibitem{qiao2013introducing}
H.~Qiao, Y.~Li, T.~Tang, and P.~Wang, ``Introducing memory and association
  mechanism into a biologically inspired visual model,'' \emph{IEEE
  transactions on cybernetics}, vol.~44, no.~9, pp. 1485--1496, 2013.

\bibitem{liu2013gnccp}
Z.-Y. Liu and H.~Qiao, ``Gnccp—graduated nonconvexityand concavity
  procedure,'' \emph{IEEE transactions on pattern analysis and machine
  intelligence}, vol.~36, no.~6, pp. 1258--1267, 2013.

\bibitem{qi2017pointnet}
C.~R. Qi, H.~Su, K.~Mo, and L.~J. Guibas, ``Pointnet: Deep learning on point
  sets for 3d classification and segmentation,'' in \emph{Proc. IEEE Conf.
  Comput. Vis. Pattern Recognit.}, 2017, pp. 652--660.

\bibitem{qi2017pointnet++}
C.~R. Qi, L.~Yi, H.~Su, and L.~J. Guibas, ``Pointnet++: Deep hierarchical
  feature learning on point sets in a metric space,'' in \emph{Proc. Adv.
  Neural Inf. Process. Syst.}, 2017, pp. 5099--5108.

\bibitem{wang2019dynamic}
Y.~Wang, Y.~Sun, Z.~Liu, S.~E. Sarma, M.~M. Bronstein, and J.~M. Solomon,
  ``Dynamic graph cnn for learning on point clouds,'' \emph{ACM Trans. Graph.},
  vol.~38, no.~5, p. 146, 2019.

\bibitem{hua2018pointwise}
B.-S. Hua, M.-K. Tran, and S.-K. Yeung, ``Pointwise convolutional neural
  networks,'' in \emph{Proc. IEEE Conf. Comput. Vis. Pattern Recognit.}, 2018,
  pp. 984--993.

\bibitem{hermosilla2018monte}
P.~Hermosilla, T.~Ritschel, P.-P. V{\'a}zquez, {\`A}.~Vinacua, and T.~Ropinski,
  ``Monte carlo convolution for learning on non-uniformly sampled point
  clouds,'' in \emph{SIGGRAPH Asia 2018 Technical Papers}.\hskip 1em plus 0.5em
  minus 0.4em\relax ACM, 2018, p. 235.

\bibitem{mao2019interpolated}
J.~Mao, X.~Wang, and H.~Li, ``Interpolated convolutional networks for 3d point
  cloud understanding,'' in \emph{Proc. IEEE Int. Conf. Comput. Vis.}, 2019,
  pp. 1578--1587.

\bibitem{hales1998overview}
T.~C. Hales, ``An overview of the kepler conjecture,'' \emph{arXiv preprint
  math/9811071}, 1998.

\bibitem{szpiro2003mathematics}
G.~Szpiro, ``Mathematics: Does the proof stack up?'' 2003.

\bibitem{hales2017formal}
T.~Hales, M.~Adams, G.~Bauer, T.~D. Dang, J.~Harrison, H.~Le~Truong,
  C.~Kaliszyk, V.~Magron, S.~McLaughlin, T.~T. Nguyen \emph{et~al.}, ``A formal
  proof of the kepler conjecture,'' in \emph{Forum of mathematics, Pi},
  vol.~5.\hskip 1em plus 0.5em minus 0.4em\relax Cambridge University Press,
  2017.

\bibitem{dai2017scannet}
A.~Dai, A.~X. Chang, M.~Savva, M.~Halber, T.~Funkhouser, and M.~Nie{\ss}ner,
  ``Scannet: Richly-annotated 3d reconstructions of indoor scenes,'' in
  \emph{Proc. IEEE Conf. Comput. Vis. Pattern Recognit.}, 2017, pp. 5828--5839.

\bibitem{roynard2018paris}
X.~Roynard, J.-E. Deschaud, and F.~Goulette, ``Paris-lille-3d: A large and
  high-quality ground-truth urban point cloud dataset for automatic
  segmentation and classification,'' \emph{Int. J. Res. Rev.}, vol.~37, no.~6,
  pp. 545--557, 2018.

\bibitem{su2015multi}
H.~Su, S.~Maji, E.~Kalogerakis, and E.~Learned-Miller, ``Multi-view
  convolutional neural networks for 3d shape recognition,'' in \emph{Proc. IEEE
  Int. Conf. Comput. Vis.}, 2015, pp. 945--953.

\bibitem{boulch2017unstructured}
A.~Boulch, B.~Le~Saux, and N.~Audebert, ``Unstructured point cloud semantic
  labeling using deep segmentation networks.'' \emph{3DOR}, vol.~2, p.~7, 2017.

\bibitem{8584494}
D.~{Lin}, R.~{Zhang}, Y.~{Ji}, P.~{Li}, and H.~{Huang}, ``Scn: Switchable
  context network for semantic segmentation of rgb-d images,'' \emph{IEEE
  Transactions on Cybernetics}, vol.~50, no.~3, pp. 1120--1131, 2020.

\bibitem{tatarchenko2018tangent}
M.~Tatarchenko, J.~Park, V.~Koltun, and Q.-Y. Zhou, ``Tangent convolutions for
  dense prediction in 3d,'' in \emph{Proc. IEEE Conf. Comput. Vis. Pattern
  Recognit.}, 2018, pp. 3887--3896.

\bibitem{ben20183dmfv}
Y.~Ben-Shabat, M.~Lindenbaum, and A.~Fischer, ``3dmfv: Three-dimensional point
  cloud classification in real-time using convolutional neural networks,''
  \emph{IEEE Robot. Autom. Lett.}, vol.~3, no.~4, pp. 3145--3152, 2018.

\bibitem{8287819}
D.~{Nie}, L.~{Wang}, E.~{Adeli}, C.~{Lao}, W.~{Lin}, and D.~{Shen}, ``3-d fully
  convolutional networks for multimodal isointense infant brain image
  segmentation,'' \emph{IEEE Transactions on Cybernetics}, vol.~49, no.~3, pp.
  1123--1136, 2019.

\bibitem{9072337}
F.~{Wang}, Y.~{Zhuang}, H.~{Zhang}, and H.~{Gu}, ``Real-time 3-d semantic scene
  parsing with lidar sensors,'' \emph{IEEE Transactions on Cybernetics}, pp.
  1--13, 2020.

\bibitem{riegler2017octnet}
G.~Riegler, A.~Osman~Ulusoy, and A.~Geiger, ``Octnet: Learning deep 3d
  representations at high resolutions,'' in \emph{Proc. IEEE Conf. Comput. Vis.
  Pattern Recognit.}, 2017, pp. 3577--3586.

\bibitem{wang2017cnn}
P.-S. Wang, Y.~Liu, Y.-X. Guo, C.-Y. Sun, and X.~Tong, ``O-cnn: Octree-based
  convolutional neural networks for 3d shape analysis,'' \emph{ACM Trans.
  Graph.}, vol.~36, no.~4, p.~72, 2017.

\bibitem{graham20183d}
B.~Graham, M.~Engelcke, and L.~van~der Maaten, ``3d semantic segmentation with
  submanifold sparse convolutional networks,'' in \emph{Proc. IEEE Conf.
  Comput. Vis. Pattern Recognit.}, 2018, pp. 9224--9232.

\bibitem{li2018so}
J.~Li, B.~M. Chen, and G.~Hee~Lee, ``So-net: Self-organizing network for point
  cloud analysis,'' in \emph{Proc. IEEE Conf. Comput. Vis. Pattern Recognit.},
  2018, pp. 9397--9406.

\bibitem{simonyan2014very}
K.~Simonyan and A.~Zisserman, ``Very deep convolutional networks for
  large-scale image recognition,'' \emph{arXiv preprint arXiv:1409.1556}, 2014.

\bibitem{he2016deep}
K.~He, X.~Zhang, S.~Ren, and J.~Sun, ``Deep residual learning for image
  recognition,'' in \emph{Proc. IEEE Conf. Comput. Vis. Pattern Recognit.},
  2016, pp. 770--778.

\bibitem{li2018pointcnn}
Y.~Li, R.~Bu, M.~Sun, W.~Wu, X.~Di, and B.~Chen, ``Pointcnn: Convolution on
  x-transformed points,'' in \emph{Proc. Adv. Neural Inf. Process. Syst.},
  2018, pp. 820--830.

\bibitem{wang2018deep}
S.~Wang, S.~Suo, W.-C. Ma, A.~Pokrovsky, and R.~Urtasun, ``Deep parametric
  continuous convolutional neural networks,'' in \emph{Proc. IEEE Conf. Comput.
  Vis. Pattern Recognit.}, 2018, pp. 2589--2597.

\bibitem{liang2019hierarchical}
Z.~Liang, M.~Yang, L.~Deng, C.~Wang, and B.~Wang, ``Hierarchical depthwise
  graph convolutional neural network for 3d semantic segmentation of point
  clouds,'' in \emph{Proc.IEEE Int. Conf. Robot. Autom. (ICRA)}.\hskip 1em plus
  0.5em minus 0.4em\relax IEEE, 2019, pp. 8152--8158.

\bibitem{landrieu2018large}
L.~Landrieu and M.~Simonovsky, ``Large-scale point cloud semantic segmentation
  with superpoint graphs,'' in \emph{Proc. IEEE Conf. Comput. Vis. Pattern
  Recognit.}, 2018, pp. 4558--4567.

\bibitem{landrieu2019point}
L.~Landrieu and M.~Boussaha, ``Point cloud oversegmentation with
  graph-structured deep metric learning,'' \emph{Proc. IEEE Conf. Comput. Vis.
  Pattern Recognit.}, 2019.

\bibitem{wang2019graph}
L.~Wang, Y.~Huang, Y.~Hou, S.~Zhang, and J.~Shan, ``Graph attention convolution
  for point cloud semantic segmentation,'' in \emph{Proc. IEEE Conf. Comput.
  Vis. Pattern Recognit.}, 2019, pp. 10\,296--10\,305.

\bibitem{jiang2019hierarchical}
L.~Jiang, H.~Zhao, S.~Liu, X.~Shen, C.-W. Fu, and J.~Jia, ``Hierarchical
  point-edge interaction network for point cloud semantic segmentation,'' in
  \emph{Proc. IEEE Int. Conf. Comput. Vis.}, 2019, pp. 10\,433--10\,441.

\bibitem{su2018splatnet}
H.~Su, V.~Jampani, D.~Sun, S.~Maji, E.~Kalogerakis, M.-H. Yang, and J.~Kautz,
  ``Splatnet: Sparse lattice networks for point cloud processing,'' in
  \emph{Proc. IEEE Conf. Comput. Vis. Pattern Recognit.}, 2018, pp. 2530--2539.

\bibitem{xu2018spidercnn}
Y.~Xu, T.~Fan, M.~Xu, L.~Zeng, and Y.~Qiao, ``Spidercnn: Deep learning on point
  sets with parameterized convolutional filters,'' in \emph{Proc. Eur. Conf.
  Comput. Vis. (ECCV)}, 2018, pp. 87--102.

\bibitem{wu2019pointconv}
W.~Wu, Z.~Qi, and L.~Fuxin, ``Pointconv: Deep convolutional networks on 3d
  point clouds,'' in \emph{Proc. IEEE Conf. Comput. Vis. Pattern Recognit.},
  2019, pp. 9621--9630.

\bibitem{komarichev2019cnn}
A.~Komarichev, Z.~Zhong, and J.~Hua, ``A-cnn: Annularly convolutional neural
  networks on point clouds,'' in \emph{Proc. IEEE Conf. Comput. Vis. Pattern
  Recognit.}, 2019, pp. 7421--7430.

\bibitem{lei2019octree}
H.~Lei, N.~Akhtar, and A.~Mian, ``Octree guided cnn with spherical kernels for
  3d point clouds,'' in \emph{Proc. IEEE Conf. Comput. Vis. Pattern Recognit.},
  2019, pp. 9631--9640.

\bibitem{zhao2019pointweb}
H.~Zhao, L.~Jiang, C.-W. Fu, and J.~Jia, ``Pointweb: Enhancing local
  neighborhood features for point cloud processing,'' in \emph{Proc. IEEE Conf.
  Comput. Vis. Pattern Recognit.}, 2019, pp. 5565--5573.

\bibitem{zhang2019shellnet}
Z.~Zhang, B.-S. Hua, and S.-K. Yeung, ``Shellnet: Efficient point cloud
  convolutional neural networks using concentric shells statistics,'' in
  \emph{Proc. IEEE Int. Conf. Comput. Vis.}, 2019, pp. 1607--1616.

\bibitem{9010002}
H.~{Thomas}, C.~R. {Qi}, J.~{Deschaud}, B.~{Marcotegui}, F.~{Goulette}, and
  L.~{Guibas}, ``Kpconv: Flexible and deformable convolution for point
  clouds,'' in \emph{2019 IEEE/CVF International Conference on Computer Vision
  (ICCV)}, 2019, pp. 6410--6419.

\bibitem{chiang2019unified}
H.-Y. Chiang, Y.-L. Lin, Y.-C. Liu, and W.~H. Hsu, ``A unified point-based
  framework for 3d segmentation,'' in \emph{2019 International Conference on 3D
  Vision (3DV)}.\hskip 1em plus 0.5em minus 0.4em\relax IEEE, 2019, pp.
  155--163.

\bibitem{narita2019panopticfusion}
G.~Narita, T.~Seno, T.~Ishikawa, and Y.~Kaji, ``Panopticfusion: Online
  volumetric semantic mapping at the level of stuff and things,'' \emph{2019
  IEEE/RSJ Int. Conf. Intell. Robots Syst. (IROS)}, pp. 4205--4212, 2019.

\bibitem{choy20194d}
C.~Choy, J.~Gwak, and S.~Savarese, ``4d spatio-temporal convnets: Minkowski
  convolutional neural networks,'' \emph{Proc. IEEE Conf. Comput. Vis. Pattern
  Recognit.}, pp. 3075--3084, 2019.

\bibitem{jaritz2019multi}
M.~Jaritz, J.~Gu, and H.~Su, ``Multi-view pointnet for 3d scene
  understanding,'' in \emph{Proc. IEEE Int. Conf. Comput. Vis. Workshops},
  2019, pp. 0--0.

\bibitem{ronneberger2015u}
O.~Ronneberger, P.~Fischer, and T.~Brox, ``U-net: Convolutional networks for
  biomedical image segmentation,'' in \emph{International Conference on Medical
  image computing and computer-assisted intervention}.\hskip 1em plus 0.5em
  minus 0.4em\relax Springer, 2015, pp. 234--241.

\bibitem{dai20183dmv}
A.~Dai and M.~Nie{\ss}ner, ``3dmv: Joint 3d-multi-view prediction for 3d
  semantic scene segmentation,'' in \emph{Proc. Eur. Conf. Comput. Vis.
  (ECCV)}, 2018, pp. 452--468.

\bibitem{huang2019texturenet}
J.~Huang, H.~Zhang, L.~Yi, T.~Funkhouser, M.~Nie{\ss}ner, and L.~J. Guibas,
  ``Texturenet: Consistent local parametrizations for learning from
  high-resolution signals on meshes,'' in \emph{Proc. IEEE Conf. Comput. Vis.
  Pattern Recognit.}, 2019, pp. 4440--4449.

\bibitem{lei2019spherical}
H.~Lei, N.~Akhtar, and A.~Mian, ``Spherical kernel for efficient graph
  convolution on 3d point clouds,'' \emph{arXiv preprint arXiv:1909.09287},
  2019.

\bibitem{thomas2018semantic}
H.~Thomas, F.~Goulette, J.-E. Deschaud, and B.~Marcotegui, ``Semantic
  classification of 3d point clouds with multiscale spherical neighborhoods,''
  in \emph{2018 International Conference on 3D Vision (3DV)}.\hskip 1em plus
  0.5em minus 0.4em\relax IEEE, 2018, pp. 390--398.

\end{thebibliography}

\begin{IEEEbiography}[{\includegraphics[width=1in,height=1.25in,clip,keepaspectratio]{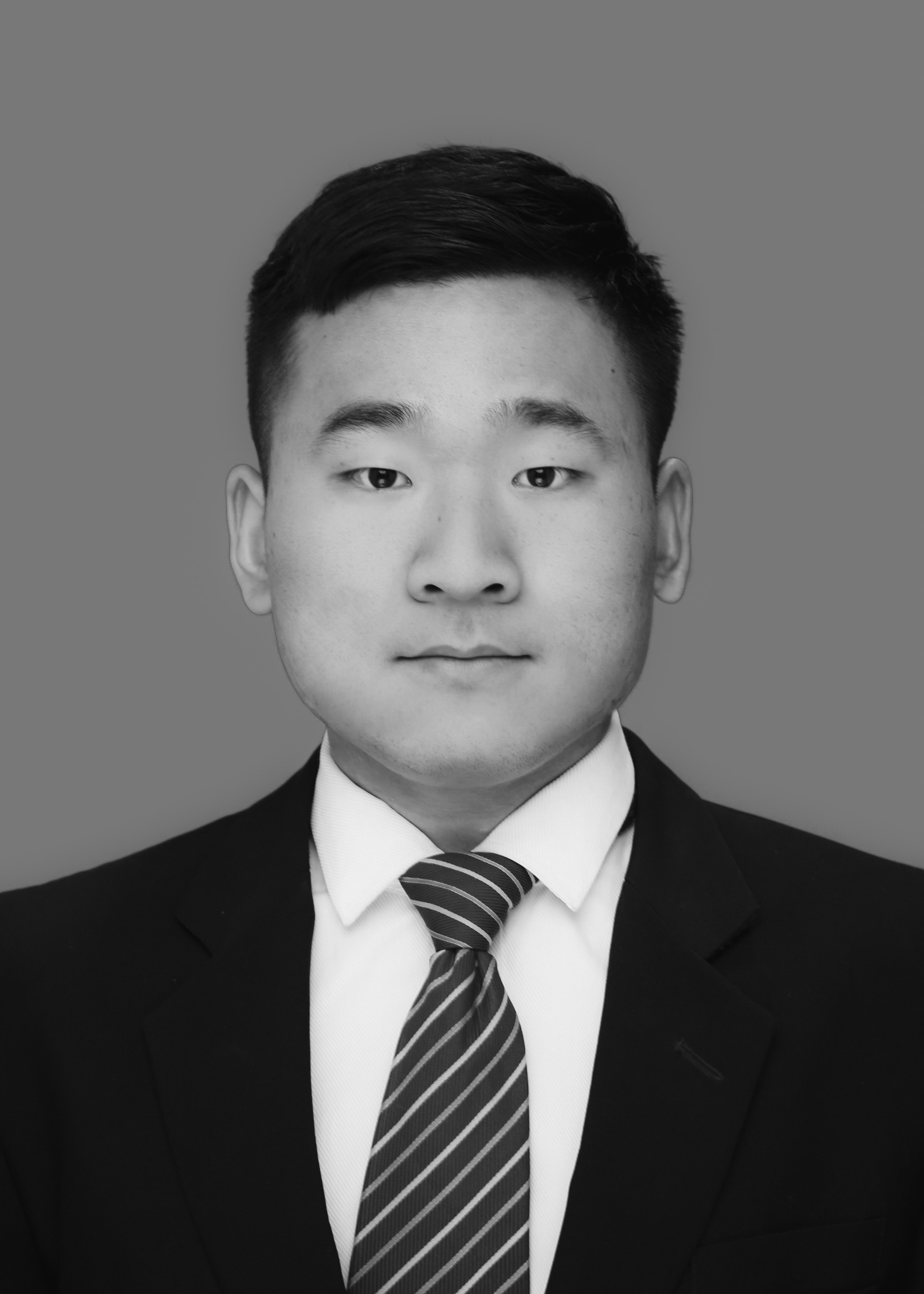}}]{Guangming Wang} received the B.S. degree from Department of Automation from Central South University, Changsha, China, in 2018. He is currently pursuing the Ph.D. degree in Control Science and Engineering with Shanghai Jiao Tong University. His current research interests include SLAM and computer vision, in particular, deep learning on point clouds.
\end{IEEEbiography}

\begin{IEEEbiography}[{\includegraphics[width=1in,height=1.25in,clip,keepaspectratio]{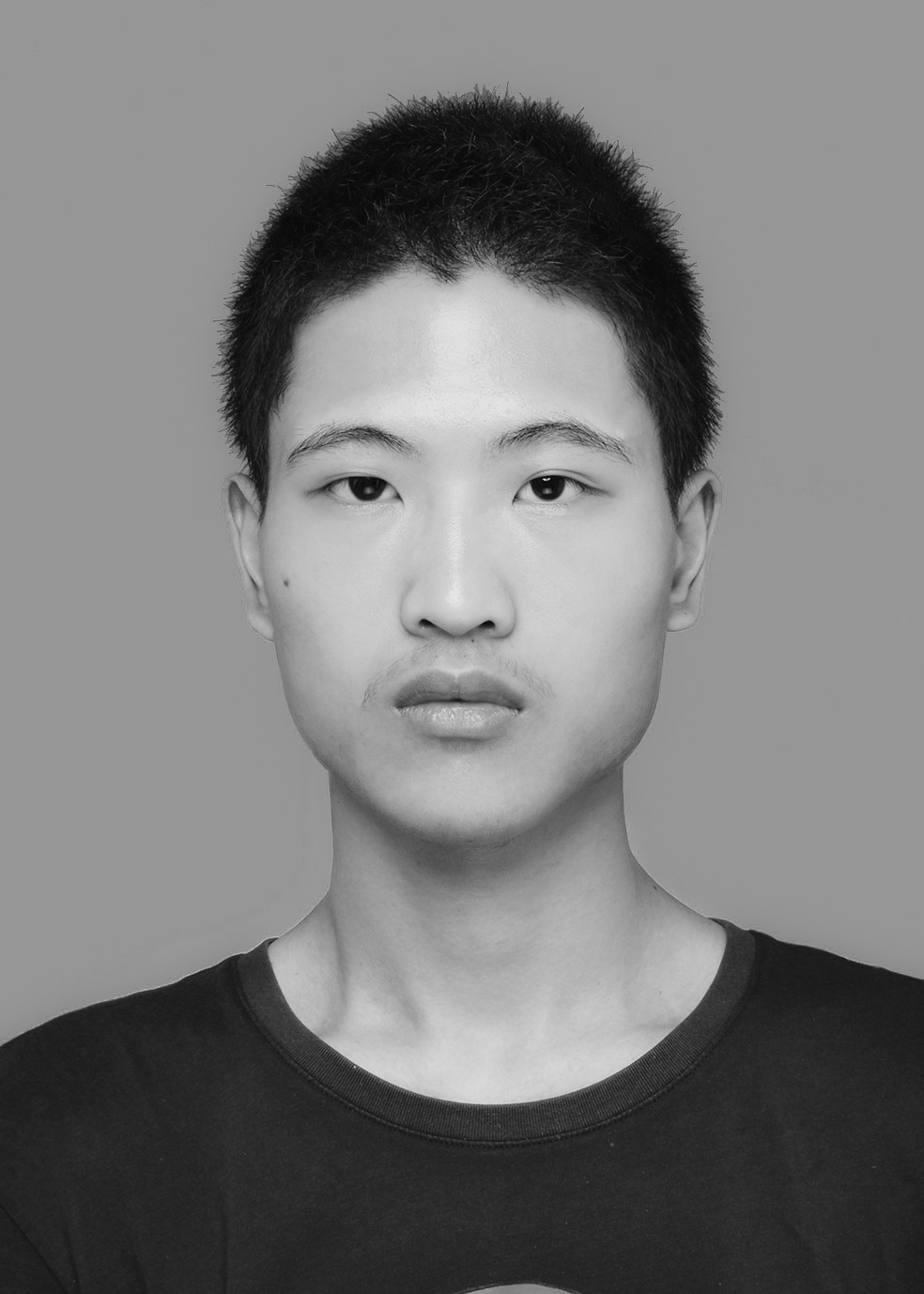}}]{Yehui Yang} 
Yehui Yang is currently pursuing the B.S. degree in Department of Automation, Shanghai Jiao Tong University. His latest research interests include SLAM and computer vision.
\end{IEEEbiography}

\begin{IEEEbiography}[{\includegraphics[width=1in,height=1.25in,clip,keepaspectratio]{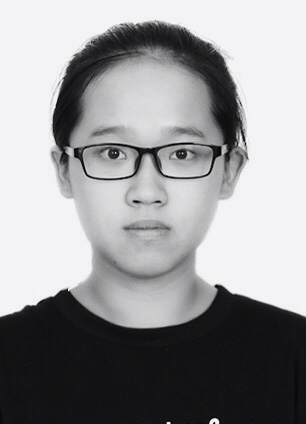}}]{Huixin Zhang} 
HUixin Zhang is currently pursuing the B.S. degree in Department of Automation, Shanghai Jiao Tong University. Her research interests include semantic segmentation for point clouds and computer vision.
\end{IEEEbiography}

\begin{IEEEbiography}[{\includegraphics[width=1in,height=1.25in,clip,keepaspectratio]{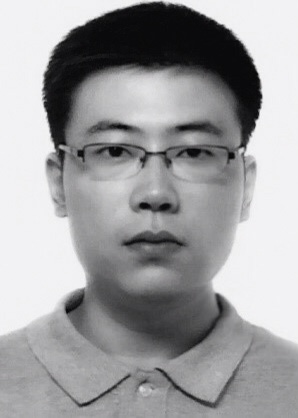}}]{Zhe Liu} received his B.S. degree in Automation from Tianjin University, Tianjin, China, in 2010, and Ph.D. degree in Control Technology and Control Engineering from Shanghai Jiao Tong University, Shanghai, China, in 2016. From 2017 to 2020, he was a Post-Doctoral Fellow with the Department of Mechanical and Automation Engineering, The Chinese University of Hong Kong, Hong Kong. He is currently a Research Associate with the Department of Computer Science and Technology, University of Cambridge. His research interests include autonomous mobile robot, multirobot cooperation and autonomous driving system.
\end{IEEEbiography}

\begin{IEEEbiography}[{\includegraphics[width=1in,height=1.25in,clip,keepaspectratio]{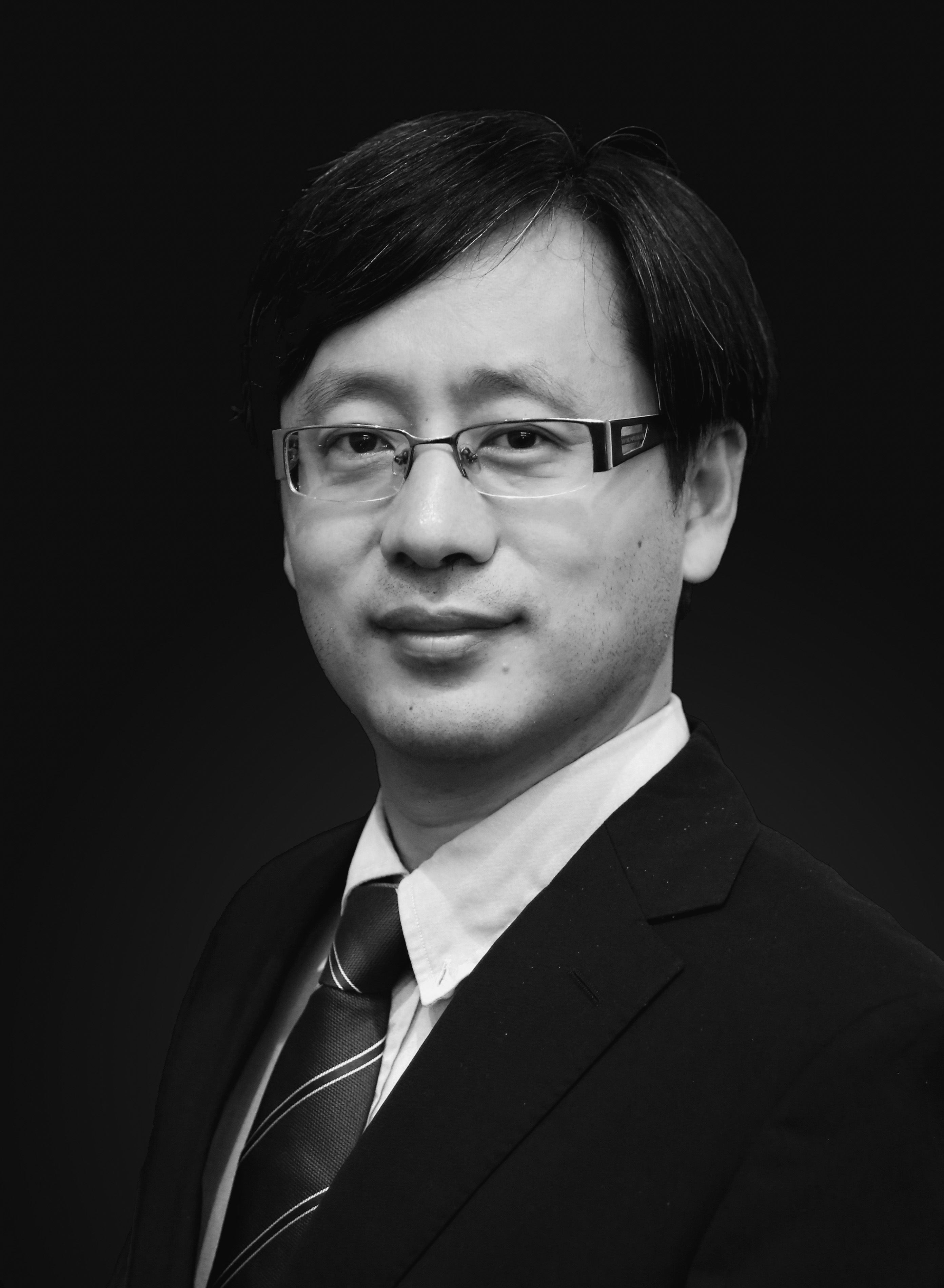}}]{Hesheng Wang} 
Hesheng Wang (SM’15) received the B.Eng. degree in electrical engineering from the Harbin Institute of Technology, Harbin, China, in 2002, and the M.Phil. and Ph.D. degrees in automation and computer-aided engineering from The Chinese University of Hong Kong, Hong Kong, in 2004 and 2007, respectively. He was a Post-Doctoral Fellow and Research Assistant with the Department of Mechanical and Automation Engineering, The Chinese University of Hong Kong, from 2007 to 2009. He is currently a Professor with the Department of Automation, Shanghai Jiao Tong University, Shanghai, China. His current research interests include visual servoing, service robot, adaptive robot control, and autonomous driving. 
Dr. Wang is an Associate Editor of Assembly Automation and the International Journal of Humanoid Robotics, a Technical Editor of the IEEE/ASME TRANSACTIONS ON MECHATRONICS. He served as an Associate Editor of the IEEE TRANSACTIONS ON ROBOTICS from 2015 to 2019. He was the General Chair of the IEEE RCAR 2016, and the Program Chair of the IEEE ROBIO 2014 and IEEE/ASME AIM 2019.

\end{IEEEbiography}

\end{document}